\documentclass{article}

\usepackage[english]{babel}

\def\etal{\emph{et al}.}
\newcommand{\vect}[1]{\boldsymbol{\mathbf{#1}}}
\usepackage{multirow,amsmath,amsfonts,amssymb,amsthm,cite}
\usepackage{newpxtext} 
\usepackage[hang,flushmargin]{footmisc}
\usepackage{placeins}

\usepackage[letterpaper,top=2cm,bottom=2cm,left=3cm,right=3cm,marginparwidth=1.75cm]{geometry}

\usepackage{amsmath}
\usepackage{graphicx}
\usepackage[colorlinks=true, allcolors=blue]{hyperref}

\title{A CNN Based Approach for the \\ Point-Light Photometric Stereo Problem}
\author{Fotios Logothetis\footnote{Toshiba Europe Ltd { flogothetis,rmecca@crl.toshiba.co.uk}}  \and   Roberto Mecca$^*$    \and Ignas Budvytis\footnote{University of Cambridge { ib255,rc10001@cam.ac.uk}}
  \and Roberto Cipolla$^\dag$
}

\begin{document}
\maketitle

\begin{abstract}

Reconstructing the 3D shape of an object using several images under different light sources is a very challenging task, especially when realistic assumptions such as light propagation and attenuation, perspective viewing geometry and specular light reflection are considered. Many of works tackling Photometric Stereo (PS) problems often relax most of the aforementioned assumptions. Especially they ignore specular reflection and global illumination effects. In this work, we propose a CNN-based approach capable of handling these realistic assumptions by leveraging recent improvements of deep neural networks for far-field Photometric Stereo and adapt them to the point light setup. We achieve this by employing an iterative procedure of point-light PS for shape estimation which has two main steps. Firstly we train a per-pixel CNN to predict surface normals from reflectance samples. Secondly, we compute the depth by integrating the normal field in order to iteratively estimate light directions and attenuation which is used to compensate the input images to compute reflectance samples for the next iteration.

Our approach sigificantly outperforms the state-of-the-art on the DiLiGenT real world dataset. Furthermore, in order to measure the performance of our approach for near-field point-light source PS data, we introduce LUCES the first real-world 'dataset for near-fieLd point light soUrCe photomEtric Stereo' of 14 objects of different materials were the effects of point light sources and perspective viewing are a lot more significant. Our approach also outperforms the competition on this dataset as well. Data and test code are available at the project page\footnote{\url{https://www.toshiba.eu/pages/eu/Cambridge-Research-Laboratory/luces}}.

\end{abstract}

\vspace{-0.01cm}
\section{Introduction}
\label{sec:introduction}
\vspace{-0.01cm}
\begin{figure*}[t]
\centering
\includegraphics[width=\textwidth]{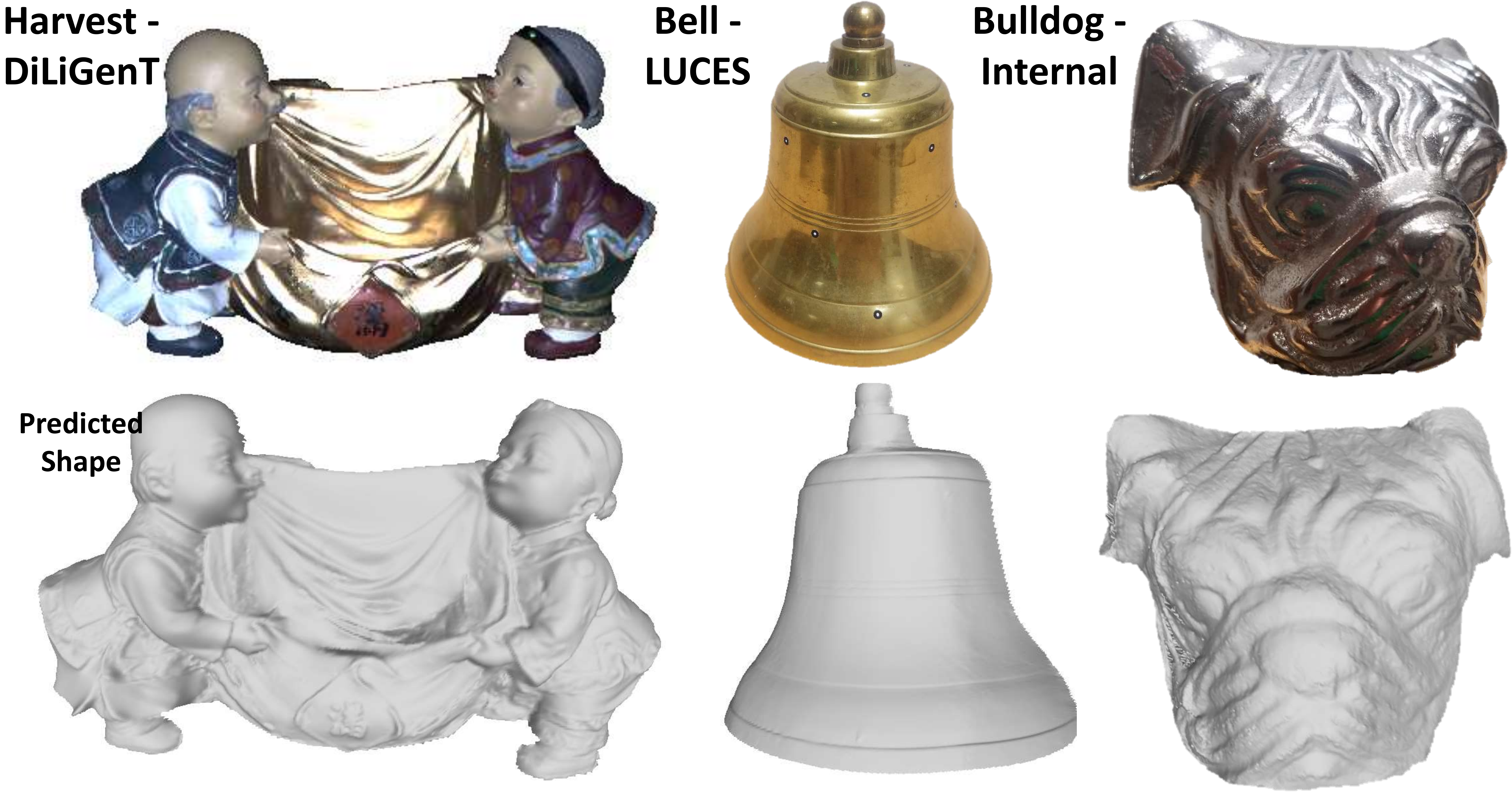}

\caption{Our proposed approach accurately reconstructs highly specular objects, in various datasets including DiLiGenT \cite{Shi2018} and LUCES \cite{mecca2021luces}.}  

\label{fig:intro}
\end{figure*}

Retrieving the 3D shape of a static object from observations under varying illumination is a very challenging problem in Computer Vision under the name of Photometric Stereo (PS). PS has been used in the past for inspection tasks such as the examination of the fracture of sandstone samples~\cite{konstantinou2021tensile} or defect detection of steel components manufacturing~\cite{SaizBAG22}.

Originally, \cite{Woodham1980} proposed a mathematical model of the PS problem relying on four main assumptions: orthographic viewing geometry, diffuse light reflection, uniform light propagation and the lack of global illumination effects (cast shadows, self reflections, ambient light). Due to restrictive assumptions, such method was limited to very narrowly specified scenarios. Since then an extensive research has been carried out to relax these assumptions. 

Shape reconstruction from shading information is a difficult problem, due to the complexity of the underlying physical process describing how a light beam bounces on the surface. Thus, it becomes very important to take into account the parametrization of all elements that influence the image formation. After \cite{Woodham1980}, most of the literature dealing with PS still assume diffuse reflection (i.e., uniform in all directions), reducing the mathematical model to a linear problem where the normal field can be easily computed \cite{Tankus2005} and finally integrated \cite{frankot1988method,queau2015edge}. Realistically, this approach contains too many assumptions which fail as soon as the method is used in a real-world application. There are at least two reasons why the reconstruction in particular of specular surfaces still remains a challenging task in the PS field. First, the Bidirectional Reflectance Distribution Function (BRDF) for specular reflections is highly non-linear, which means that analytical solutions are intractable. 
Second, the behaviour of light before and after it starts bouncing on the object needs to be modeled accurately. As many recent methods \cite{ikehata2018cnn, logothetis2021pxnet, santo2020deep} aim at retrieving the geometry at a per-pixel level, light-object interaction requires proper modeling.

With the aim of solving the PS problem under more realistic conditions, researchers have modelled perspective viewing geometry \cite{Prados2003, Tankus2005,Onn1990}, specular light reflections \cite{MeccaQLC2016} 
and point light sources parameterising radial propagation of light \cite{Iwahori1990point,Clark1992}. These effects lead to highly non-linear models requiring sophisticated optimisation strategies \cite{WMBK14,queau2018led}. As the complexity of the models becomes intractable, especially when dealing with physical material properties, several recent works have opted to neglect specular highlights and instead rely on robust optimisation techniques \cite{Ikehata2012Robust,6909677}. Furthermore, real objects experience a number of complex physical effects which make the explicit mathematical modeling very hard to invert. 

\begin{figure*}[t]
 \includegraphics[width=\textwidth]{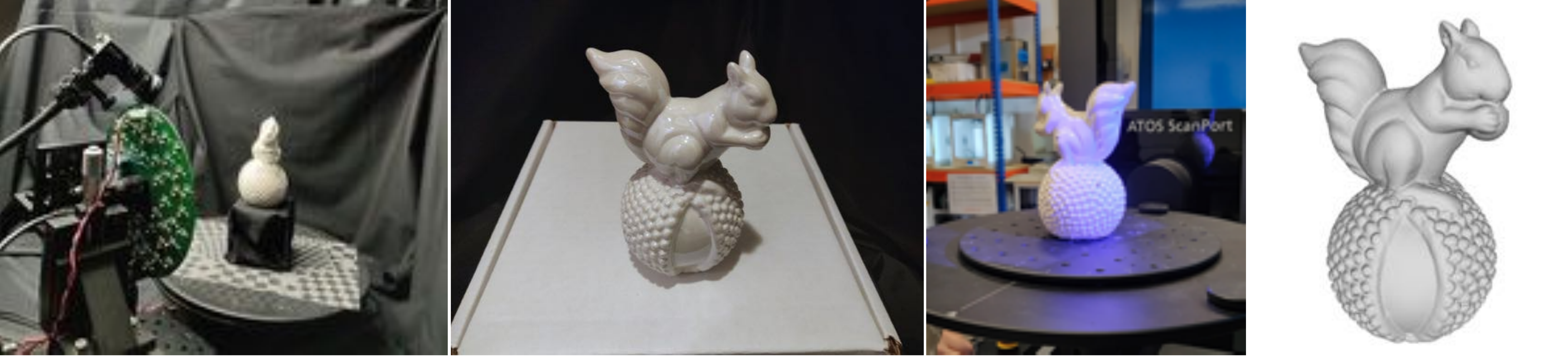}
 \caption{From left to right: (1) the stage of our Photometric Stereo setup, (2) a top view of a sample object (Squirrel), (3) acquisition with the GOM scanner, (4) the 3D scanned mesh.}
 \label{fig:setup}
\end{figure*}

In fact, these global illumination effects (cast shadows, self reflections, ambient light) are one of the most challenging aspects of PS. \cite{logothetis2016near,YuilleSEB99} tackle the case of fixed ambient light which however is too simple of a model to cover realistic inter-reflections. The global illumination issue is firstly adequately addressed in~\cite{ikehata2018cnn} by employing a Convolutional Neural Network (CNN). 
This method works by arranging reflectance samples into a fixed size observation map for each pixel. Observational maps are then provided as an input to the CNN which is trained to output a surface normal per pixel. \cite{logothetis2021pxnet} extends this work and shows how a training data augmentation strategy can be used to deal with general BRDFs such as MERL dataset~\cite{Matusik03} or Bidirectional Scattering Distribution Functions (BSDF) such as Disney~\cite{burley2012physically} and global illumination effects in the far-field setting. However, these approaches are only directly applicable to the far-field photometric stereo since the nonlinear light attenuation of near-field images does not allow to directly compute valid observation maps.  

Usually the concept of near-field PS is relevant when the images are acquired with the camera/light setup nearby the object. Differently from the far-field case where incoming light is parametrised as a uniform 3D vector, 
light directions at every pixel location are dependent on the geometry of the light source. Instead of dealing with general lighting models ~\cite{SPIC2016}, most of the approaches consider a point-like light source which actually matches the widely available LED based illumination. It is important to notice that even far-field PS datasets are acquired by using point light sources~\cite{diligentshi2016benchmark}. In this work we use the concept of point-light based PS and, instead of constraining it in the near-field, we provide a method which is able to improve state-of-the-art also in the far-field.

To do so, we use a three step process. Firstly, the effect of the light attenuation is compensated using an estimate of the object geometry, to produce equivalent far-field reflectance samples. Secondly, a CNN is used to regress pixel normals from these samples. Finally, a numerical integration is used to update the estimate of the object geometry for the next iteration step. We evaluate our method on both artificial and real point-light image datasets. We significantly outperform competing approaches~\cite{queau2018led,logothetis2017semi,santo2020deep} on both types of datasets 
(see Figure~\ref{fig:intro} and Section~\ref{sec:experiments}).

We extend our previous approach \cite{logothetis2020bmvc} by making the network able to train over a general point-light distribution. We tested over a wide variety of scenarios, taking into account sparse and dense point-light distribution as well as synthetic experiments. We finally compare our method over the real-world point-light PS datasets DiLiGenT~\cite{diligentshi2016benchmark} and LUCES~\cite{mecca2021luces} to cover both far and near field setting respectively. We also extend the preliminary version of LUCES (see Figures~\ref{fig:setup},~\ref{fig:ndsetup} and Section~\ref{sec:LUCES}) \cite{mecca2021luces} by analysing the real to synthetic gap among a variety of competing methods. In addition, we improved ground truth meshes by employing CT scanning technologies\footnote{\url{www.zeiss.com/metrology}} to retrieve 3D geometry of objects made of non-diffuse materials. Under different lighting setups between LUCES and DiLiGenT, object materials, focal length and illumination density, we discuss the variation of performances for different configuration of several approaches.

The rest of this work is divided as follows. Section 2 discusses relevant work in Photometric Stereo. Section~\ref{sec:method} provides details of our proposed method. Section~\ref{sec:LUCES} outlines the LUCES dataset. Sections~\ref{sec:experimental_setup} and \ref{sec:experiments} describe the experiment setup and corresponding results.


\section{Related Work}
\label{sec:relatedWorks}

In this section we provide an overview of the relevant latest improvements in PS. 
For a detailed, fairly recent PS survey, refer to \cite{AckermannG15}. 

\subsection{Point-light Source PS} Differently from the classical directional-light PS, point-light source based approaches assume that the illumination spreads non-linearly with respect to the position of the light sources, thus making analytical models more complicated and harder to solve in practice. 

Most of the approaches that dealt with point-light illumination were actually trying to solve specific applications mostly related to endoscopic inspection \cite{Deguchi,Wu2010Endoscope,Collins2012,Parot2013}. In particular they were always trying to tackle the problem under near-field setting, which is the most obvious scenario where non-linear light behaviour and image perspective have to be addressed in order to avoid distorted geometry.

In this particular endoscopic framework, Wu \etal \cite{Wu2010Endoscope} studied the multi-image endoscopic problem by considering two light sources placed off the optical center. They developed an irradiance model obtained by simultaneously illuminating an object with two different light sources. They then recovered the surface by considering a single irradiance equation for the sum of Lambertian reflectance functions of the two different light sources. The use of this reflectance function results in a loss of information. In order to avoid this problem and issues related to an unknown albedo, they used a photometric calibration. Surface recovery is performed within a variational framework that involves high computational complexity compared to alternative direct methods \cite{MeccaFalcone2013}. The shape from an endoscopic perspective problem solved via a photometric stereo technique using more than 2 images was first addressed by Collins and Bartoli \cite{Collins2012}. They solved the close-range PS with with an a-priori light calibration procedure. Furthermore, they used a prior for a reflectance model learning by adding physical markers on the inspected object even when the surface was assumed to be Lambertian. In particular, their mathematical formulation is based on the usual two step procedure where an energy functional is minimized (which allows the computation of the surface derivatives), and only later is the surface recovered \cite{Agrawal2006,frankot1988method,simchony1990direct}. Moreover, their energy is based on the sum of Lambertian irradiance equations rather than using photometric ratios \cite{Wolff1994ratio,MeccaFalcone2013,ChandrakerAK07,VogiatzisH10} that lead to more practical problems. For example, the most important feature of photometric ratios is to obtain independence from the albedo. Parot \etal \cite{Parot2013} studied the same problem by using a straightforward heuristic approach to photometric stereo. In their work, even if camera and lights are close to the inspected object, they assumed orthographic viewing geometry, with uniform and unattenuated light directions calibrated by assuming reasonable distance between the object and the camera. The discrepancy with respect to the real physics about object proximity is faced by filtering the directional gradients depending on the frequencies. They heuristically handled this by removing the lower frequencies and the DC components. Then, the resulting depth map is computed using a multigrid Poisson solver \cite{simchony1990direct}. The work describes purely qualitative results in the sense that they did not represent accurate reconstructions of the environment, instead they used their method as a qualitative tool for detecting lesions.

Some works embedded the non-linearities coming from the point-light source geometry in a Partial Differential Equation (PDE)-based formulation using image ratios \cite{Mecca2014near,MeccaRC15}. This way allows to calculate depth directly, without the intermediate step of approximating the normal field. 
Also \cite{Lee1991glaucoma,SmithFang2016} took advantage of image ratios in order to eliminate the dependence on the surface albedo and thus reduce the number of unknowns. Image ratios were also used in the variational framework of \cite{MeccaQLC2016} in order to make the approach more robust to specular highlights by unifying diffuse and Blinn-Phong specular \cite{Blinn:1977} reflections into a single mathematical formulation. This general variational framework is also applicable in a weakly calibrated setting \cite{logothetis2017semi} or even a volumetric one \cite{logothetis2019differential}. 
Recently, a LED-based approach introduced by Qu\'eau \etal ~\cite{queau2018led} presented a complicated variational approach based on alternating weighted least-square scheme also capable of calibrating the light brightness of the light sources. Furthermore, \cite{liu2018near} exploited a circular LED setup to compute the relative mean distance between the camera and the object.

\subsection{Deep Learning (DL) Based Approaches for PS} 
Computer graphics is a well understood topic and many tools capable of rendering highly non-linear irradiance equations are publicly available\footnote{\url{www.blender.org} and \url{www.disneyanimation.com/technology/brdf.html}} \cite{Matusik03}.
This allowed to create reliable datasets for supervised DL approaches. The potential of DL for solving the PS problem can be divided in two main advantages. Firstly, CNNs have the capability of inverting highly non-linear reflectance models comprising of numerous physically based parameters. Secondly, CNNs can be made to deal with the complicated real world imperfection (shadows, self reflections, noise) through the use of data augmentation.
So far, several DL approaches have been proposed \cite{ikehata2018cnn,logothetis2021pxnet, santodeep,santo2020deep}. 
A preliminary work by \cite{TangSH12a, Hinton09} 
considered diffuse reflection only. \cite{YuS17} 
proposed a library where set of novel layers can be incorporated into a generic neural network to embed explicit models of photometric image formation. More recently, several approaches have tackled the problem of reconstructing complex objects. \cite{santodeep} proposed a method to find correspondences between simulated observation rendered by the MERL BRDF dataset \cite{Matusik03} and the normal map of the target object,  handling non-local effects using a dropout strategy.
\cite{JuQZDL18} leveraged DL to learn the information from multispectral images to get RGB based PS reconstructions. 
\cite{taniai2018neural} proposed generating training data on the go to minimise the image re-projection error. Although this method is a training data free approach, the whole procedure is relatively slow.  \cite{li2018learning} proposed a dedicated network to account for global illumination effects for the case of single mobile image reconstruction. Recently, \cite{ChenHW18} proposed rendering patches of different surface materials in order to get training data. This method is also extended in \cite{chen2019self} for solving the uncalibrated PS. 
Ikehata \cite{ikehata2018cnn} proposed arranging all the reflectance samples of a pixel (i.e. different illumination images in the far-field setting) into a fixed size \textit{observation} map which is used by a CNN to regress pixel normals. The CNN is essentially learning to invert the BRDF with added robustness to global illumination effects, as training data are made with physics based rendering. In  \cite{logothetis2021pxnet}, this method was extended by simplifying the training procedure providing an inline per-pixel training data generation.

However, none of these DL approaches directly tackle the point-light PS problem and non-linear attenuation from point light sources together with the viewing direction dependency drastically increase the problem space exploding the training data requirements. Santo \etal \cite{santo2020deep} addressed this problem with a hybrid approach where the light reflected is firstly interpreted as coming from a directional light source, and then refined with a point-light model based on a near-light image formation.

In this work, we expand our method \cite{logothetis2020bmvc} which was limited to provide depth prediction for the point-light PS problem for a specific light configuration. The proposed per-pixel training procedure has been improved in order to include a much wider variety of lighting scenarios. This allows the proposed network to provide state-of-the-art predictions in a general point-light setup.

\begin{figure*}[ht]
\centering
\includegraphics[width=\columnwidth]{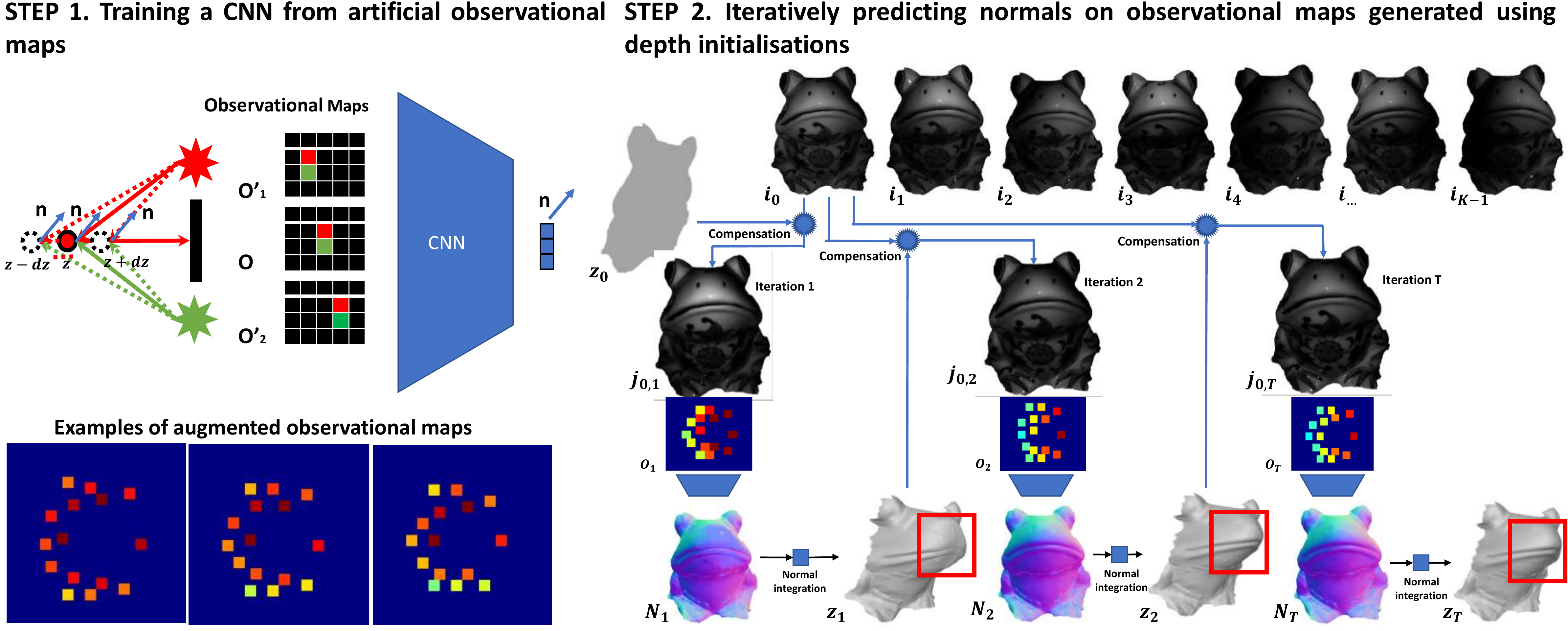}
\vspace{-0.25cm}
\caption{This figure illustrates two key steps of our proposed approach. On the left, the network training is illustrated consisting of sampling points inside the camera's frustrum and rendering the respective observation maps. As the depth will only be approximately known at test time, this is slightly perturbed before mapping resulting to a structured change of the map (this structured change is shown at bottom left: the middle image is the map computed with the actual depth (10~cm), the left and right maps are computed with 9 and 11 cm respectively). On the right, the reconstruction process is shown. Images $i_{0}\cdots i_{K-1}$ are used with conjunction with previous depth estimate to compensate for light attenuation ($j_{0}\cdots j_{K-1}$), compute observation maps (shown for pixel at image center here), regress normals and finally update the shape. Note the improvement of the shape of frogs beak (red square) from iteration 1 to iteration 2. Also see Fig.\ref{fig:normal_pred}.} 
\label{fig:method}
\end{figure*}
\begin{figure*}[ht]
\centering
\includegraphics[width=\textwidth]{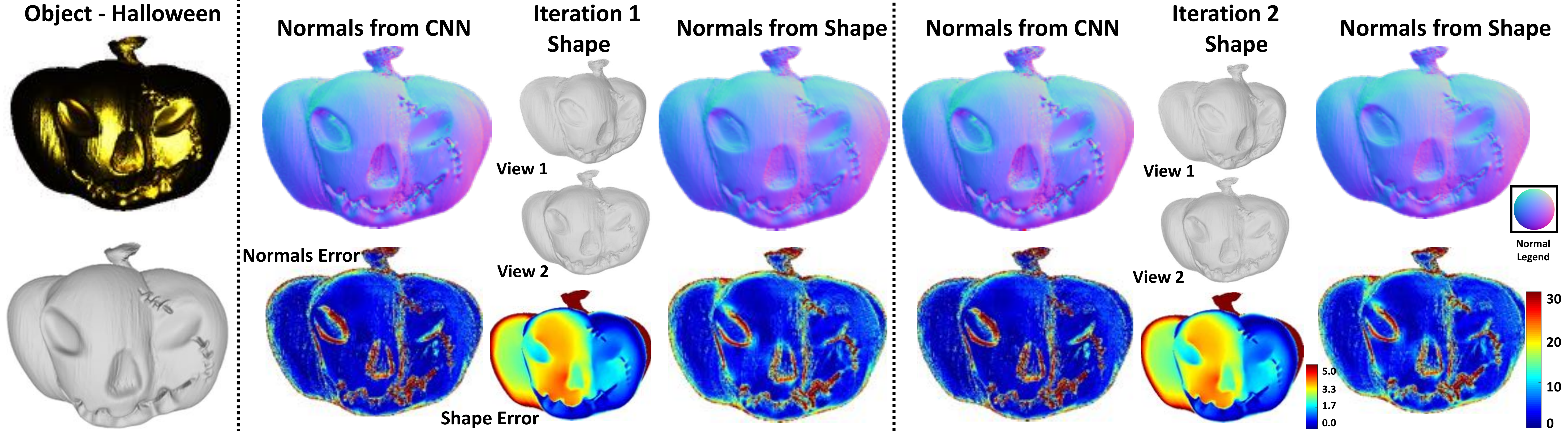}
\vspace{-0.25cm}
\caption{Iterative refinement of the geometry for the Halloween synthetic object. On the left, a sample image and GT shape are shown. The other 2 sections show 2 steps of the iterative refinement with the respective normals (both raw network predictions and differentiated ones), normal error maps and depth error maps. As the difference is minimal between steps 1 and 2, the process is converged.}
\label{fig:normal_pred}
\end{figure*}

\subsection{Photometric Stereo Datasets} 

Across the years, a number of custom real-world PS datasets have been created to suit the purposes of the proposed approaches. Alldrin \etal \cite{Alldrin2008} proposed a dataset consisting of 3 objects lit by roughly a hundred distant directional lights. The light calibration in terms of positioning and intensity has been performed by using respectively a mirror sphere and a diffuse sphere. Xiong \etal \cite{XiongCBGJZ15} have proposed a dataset of 7 objects using 20 directional lights calibrated with two chrome spheres. As the approach was mostly modeling PS images with Lambertian irradiance equations, the material of the objects was quite diffuse. A limited number of PS data has been released by Qu\'eau \etal ~to prove the working principle of an edge preserving method \cite{queau2015edge} and a multi-spectral PS approach \cite{Queau2016colored}.

Although initially designed for evaluating multi-view approaches, the datasets released by Aan{\ae}s \etal \cite{AanaesDP12, AanaesJVTD16} are useful for evaluating PS approaches as they also contain images under varying illumination.
As most of the methods aimed at tackling the PS problem deal with the far-field setting, recently Shi \etal \cite{Shi2018} introduced the first dataset in this category, namely DiLiGenT aimed at evaluating reconstruction methods over a wide variety of materials for 10 different objects. This work also contains a well discussed taxonomy for non-Lambertian and uncalibrated PS approaches. Their setup consists of 96 LEDs placed several meters away from the objects to approximate directional illumination and the camera
(with a 50mm lens) was placed at 1.5m from the object. Such distance between the object and the camera/lights system does not provide to this dataset the near-field light variation studied in many recent approaches.


\section{Method}
\label{sec:method}

In this section we describe our method for tackling the point-light Photometric Stereo problem. In particular, we provide both the details of the assumed image formation model and how normals can be predicted for Photometric Stereo images by using CNN's trained on reflectance samples (also see Figure~\ref{fig:method}).

\subsection{Point-light Modeling}
\label{sec:method_point}
Similar to \cite{Mecca2014near}, we assume calibrated point light sources at positions $\mathbf{P}_m$ (w.r.t the camera center at $\mathbf{0}$) resulting in variable lighting vectors  $\mathbf{L}_m=\mathbf{P}_m-\mathbf{X}$. Here $\mathbf{X}=[x,y,z]^ \intercal$ is the 3D surface point coordinates. 
We also model the light attenuation considering the following non-linear radial model of dissipation:
\begin{equation}
\label{eq:attenuation}
a_m(\mathbf{X})=\phi _m \frac{( \hat{\mathbf{L}}_m (\mathbf{X}) \cdot \hat{\mathbf{D}}_m)^{\mu _m}}{||\mathbf{L}_m (\mathbf{X})||^2},
\end{equation}
where $\hat{\mathbf{L}}_m=\frac{\mathbf{L}_m}{||\mathbf{L}_m||}$ is the lighting direction, $\phi _m$ is the intrinsic brightness of the light source,~$\hat{\mathbf{D}}_m$ is the principal direction (i.e. the orientation of the LED point light source) and $\mu _m$ is an angular dissipation factor. Defining $\hat{\mathbf{V}}=-\frac{\mathbf{X}}{||\mathbf{X}||}$ as the viewing vector, the general image irradiance equation becomes: 
\begin{equation}
\label{eq:irad}
i_m= a_m \text{B}(\mathbf{N},\hat{\mathbf{L}}_m,\hat{\mathbf{V}},\rho).
\end{equation}
Here $\mathbf{N}$ is the surface normal. B is assumed to be a general BRDF and $\rho$ is the surface albedo (allowing for the most general case, images and $\rho$ are RGB and the reflectance is different per channel). In addition, we allow for the possibility of global illumination effects (shadows, self reflections) which are incorporated into B. This can be re-arranged into a BRDF inversion problem as (for BRDF samples $j_m$):
\begin{equation}
\label{eq:brdf_inv}
j_m=\frac{i_m}{ a_m}= \text{B}(\mathbf{N},\hat{\mathbf{L}}_m,\hat{\mathbf{V}},\rho).
\end{equation}

\noindent
We note that $\hat{\mathbf{V}}$ is known but $\mathbf{L}_m$ and $a_m$ are unknowns due to the nonlinear dependence on $z$. Our objective is to recover the surface normals $\mathbf{N}$ and depth $z$. 

\subsection{Normal Prediction} The first step of our method includes training a CNN for per-pixel normal prediction using BRDF samples. This is done through the the observational map parameterisation introduced by \cite{ikehata2018cnn} in order to tackle the far-field photometric stereo problem. Note that this is equivalent to BRDF inversion under the special case of  $\hat{\mathbf{V}}=[0,0,1]$.
As described in~\cite{ikehata2018cnn} an observational map records relative pixel intensities (BRDF samples) on a 2D grid (e.g. $32 \times 32$) of discretised light directions. Such a representation is highly convenient for use with classical CNN architectures as it provides a 2D input and is of fixed shape despite a potentially varying number of lights. While~\cite{ikehata2018cnn} proposes to train CNNs on rendered images of objects, it is shown in~\cite{logothetis2021pxnet} that simpler per-pixel renderers can be used instead, making the training procedure much faster and simpler. We use the latter approach in this work. Following ~\cite{logothetis2021pxnet}, an RGB observation map $O_\text{rgb}$ of size $d \times d \times 3$ is constructed as:
\begin{equation}
O_\text{rgb}\Big( \Big \lfloor d\frac{\hat{L}^x_m+1}{2}  \Big \rfloor, \Big \lfloor d\frac{\hat{L}^y_m+1}{2}  \Big \rfloor \Big)=
\begin{bmatrix}
j_{\text{r}}/\phi _{\text{r}} \\
j_{\text{g}}/\phi _{\text{g}} \\
j_{\text{b}}/\phi _{\text{b}} \\
\end{bmatrix}_m.
\label{eq:omaprgb} 
\end{equation}

In addition, we note that in the case of specular reflection, the BRDF samples $j$ are dependent on the viewing vector $\vect{V}$. This variation is only expected to be significant in the case of perspective projection for points not close to the imaging center. Nonetheless, the set of orthographically rendered observation maps considered in \cite{ikehata2018cnn} or \cite{logothetis2021pxnet} is only a special case of the possible observation maps. Thus, to make the network training problem easier, we extend the observation map concept to incorporated the viewing vector  $\vect{V}$ (which is known and constant for all light sources $m$) such as: 
\begin{equation}
 O=[O_\text{rgb} ~;  \vect{1} \vect{V}]
\label{eq:omapall} 
\end{equation}
where $ \vect{1}$ is $d \times d \times 3$ and ;  is a  concatenation on the 3rd axis so defining a $d \times d \times 6$ map. Finally, these observation maps are fed into a CNN which regresses surface normal $\vect{N}_p$. The CNN is trained with the angular loss defined as  $|\text{atan2}(||\vect{N}_t \times \vect{N}_p||, \vect{N}_t \cdot \vect{N}_p)|$ with $\vect{N}_p$ are the predicted normals and $\vect{N}_t$ are the ground truth normals .

\subsection{Adapting to the Point Light Setup}
In order to solve the point light PS problem for a realistic capture setup 
we adapt the training procedure to only sample observation maps which are plausible at test time. Therefore, instead of sampling a random set of light directions as in \cite{logothetis2021pxnet}, we sample 3D points inside a virtual camera frustum. For each point, a different LED configuration is simulated.

\begin{figure*}[t]
\centering
\includegraphics[width=0.95\textwidth]{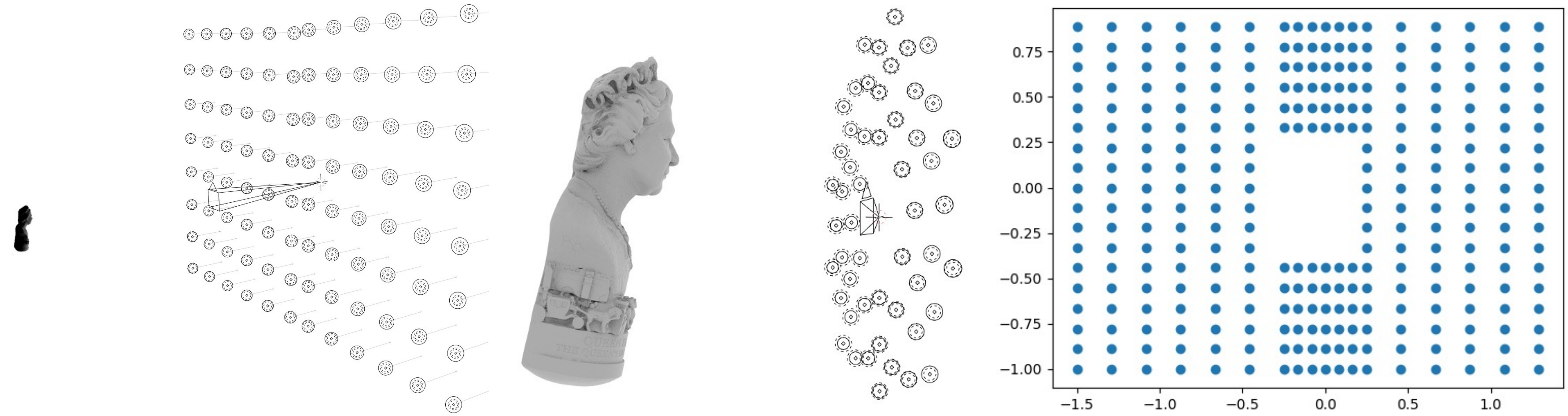}
\caption{On the left the disposition of point-lights for the DiLiGenT dataset \cite{diligentshi2016benchmark}. On the middle the one for the LUCES dataset \cite{mecca2021luces}. In order to give idea of their scale, both configurations have been rendered facing the same object Queen at exact distance/orientation from the camera/light setup used for generating the PS images in the respective datasets (real LUCES and synthetic DiLiGenT). On the right, a potential (out of the ones sampled at train time) point light configuration is shown. This is a rectangle of sides 3x2 with a hole in the middle of size 0.33x0.66 (sizing is in terms of $z$) containing the maximum of 288 lights.} 
\label{fig:ndsetup}
\end{figure*}

\noindent
\textbf{Configuration sampling}. 
The sampling procedure for a point begins with sampling normalised image plane coordinates $[u,v] \in [-1,1]$. Then camera focal length $f \in[1,10]$ is sampled.  $f=1$ corresponds to a real fish eye lens and $f=10$ is close to orthographic viewing. For reference LUCES normalised focal length is $ \approx 1.5$, DiLiGenT is $ \approx 5$. Then a depth $z$ is sampled in a range from 10cm to 170cm and this depth is used to back-project image coordinates and obtain 3D point in camera coordinate system $\vect{X}=[u z/f, v z /f,z]^T$. 

The rest of the configuration is sampled proportionally to $z$ which allows\footnote{We assume that in most cases the radius of the arrangements of point lights would be of similar scale as the width of the objects which is being scanned.} for tackling LED arrangement of vastly different scales (see Figure~\ref{fig:ndsetup}). We assumed that point lights are approximately on a plane parallel to $XY$ axes. The plane offset is uniformly sampled in the range $ [0, 0.25z]$ and all lights are positioned at a height with respect to that plane up to $\pm 0.05z$. In terms of distribution of the lights on that  plane, we assume a rectangle with side lengths in $[0.5z , 3z]$ and with a rectangular hole in the middle with side lengths $[0, 0.66z]$  (see Figure~\ref{fig:ndsetup}). This plane area is divided into a grid and a number between 15 and 288 points of this grid are selected to be the light positions  $\mathbf{P}_m$. Light brightness $\phi _m$, are sampled uniformly and independently in log scale from $\phi _m \in [0.25,4]$, dissipation factors $\mu \in[0,3]$  and $\mathbf{D}_m = [d_x,d_y,1+d_z] \text{~with~} d_{x,y,z} \in [-0.1,0.1]$ (ensuring $||\mathbf{D}_m||=1$). Finally, surface normal $\vect{N}$, material parameters and global illumination approximations are sampled independently following the exact same hyper-parameters of \cite{logothetis2021pxnet}.

\noindent
\textbf{Reflectance rendering}. Once the point parameters are sampled, the image samples $i_m$ are rendered  (Equation~\ref{eq:irad}) using the renderer of \cite{logothetis2021pxnet} with the addition of point light attenuation (Equation~\ref{eq:attenuation}). Additional global illumination approximations for shadows, reflections and ambient light are also applied as in~\cite{logothetis2021pxnet}. Schematically this corresponds to:
\begin{equation}
\label{eq:redner_fw}
\{ \vect{X}, \mathbf{P}_m,  \phi _m, \mathbf{D}_m, \mu _m, \vect{N} \} \xrightarrow {\text{Eq.\ref{eq:attenuation}}} \{ \hat{\mathbf{L}}_m, a _m\} \xrightarrow {\vect{N},\text{\scriptsize Render}} \{i_{m} \}. 
\end{equation} 
\noindent
Note rendered intensities $i_m$ are rendered using constant light attenuation $a _m=\phi _m$. The final intensities $i_m$ are obtained by using non-linear radial model of dissipation described in Eq.~\ref{eq:attenuation}. Also note that 10bit discretisation, and saturation are applied (i.e. conversion to integer $\in \{0,1023\}$ and re-normalisation) when rendering values $\{i_m \}$ 
to approximate the camera of LUCES.

\noindent 
\textbf{Observation map generation}. After performing the point rendering to compute  $i_m$, the aim is to compensate for light attenuation to compute reflectance sample (Equation~\ref{eq:brdf_inv}) and generate observation maps  (Equation~\ref{eq:omaprgb}). In order to get robustness to imprecise depth initialisation at test time, the training procedure involves perturbing the ground truth depth value $z$  by $\delta z\sim \mathcal{N}(0,5\%z)$\footnote{The Gaussian distribution encourages the network to be more accurate when $\delta z$ is small and thus get an improvement in the iterative setting.} to obtain $z'=z+\delta z$. In addition, all setup parameters are also slightly perturbed to account for potential setup miss-calibration i.e.:
\begin{equation}
\label{eq:pertudbation}
\{ z, \vect{X}, \mathbf{P}_m,  \phi _m, \mathbf{D}_m \mu _m \} \xrightarrow {\delta}  \{  z', \vect{X'}, \mathbf{P'}_m,  \phi ' _m, \mathbf{D}'_m \mu ' _m \} 
\end{equation}
The hyper-parameters in Equation~\ref{eq:pertudbation} are: \\ $ \delta \mathbf{P} \in \left[-0.1\%z,0.1\%z\right]$ (additive),  
$ \delta \phi \in [0,1\%$] (multiplicative), $\delta \mathbf{D} \in [-0.1,0.1]$ (additive),
 $\delta \mu_{1} \in [0,0.1]$ (additive) and   $\delta \mu_{2} \in [0,10\%]$ (multiplicative). We note that we samples these perturbations both independently for all light sources but also include an additional amount of perturbation (of same distribution) to all the lights at the same time to account for systematic errors. Finally, these perturbed parameters are used to recompute attenuation, reflectance samples and then observation maps $O'$ i.e.:
\begin{equation}
\label{eq:maps_after_pertubation}
\{\vect{X'}, \mathbf{P'}_m,  \phi ' _m, \mathbf{D'}_m \mu '_m \} \xrightarrow {\text{Eq.\ref{eq:attenuation}}}  \{  \hat{\mathbf{L'}}_m, a' _m \} \xrightarrow {i_m,~ \text{Eq.\ref{eq:brdf_inv}-\ref{eq:omapall}}} O'
\end{equation}


\noindent
\textbf{Specific setup}. In the case of aiming to train a network for a specific dataset (e.g LUCES~\cite{mecca2021luces}) the plausible observation map space is highly reduced since the camera/light configuration is known, and thus the data generation process can take advantage of that. We note that setup knowledge means that the parameters used to compute the observation maps at test time, i.e $\{\mathbf{P'}_m,  \phi ' _m, \mathbf{D'}_m \mu '_m \}$ (in Equation~\ref{eq:maps_after_pertubation}) are assumed to be known at train time. Of course, it is still desirable to have some robustness to potential setup miss-calibration, therefore the perturbation equation is applied in \textit{reverse} order (so as to end up at the map creation stage with the parameters that will be used at test time). This is summarised as:

\begin{align}
\label{eq:pertudbation:spec}
&\text{Calibration} \xrightarrow {\text{Copy}}  \{ \mathbf{P'}_m,  \phi ' _m, \mathbf{D}'_m \mu ' _m \} \\
& \{\mathbf{P'}_m,  \phi ' _m, \mathbf{D}'_m \mu ' _m \}   \xrightarrow {\delta} \{\mathbf{P}_m,  \phi _m, \mathbf{D}_m \mu _m \}  \\
& \text{Eq. \ref{eq:redner_fw}, \ref{eq:maps_after_pertubation}}  \rightarrow O'
\end{align}

\subsection{Iterative Refinement of Depth and Normals} 
Assuming an estimate of normals, the depth can be obtained by numerical integration. This is performed using the $\ell _1$ method of \cite{queau2015edge}. The variational optimisation includes a Tikhonov regulariser with $z=z_0$ ($z_0$ is the previous estimate of the depth map starting from plane) weight $\lambda =10^{-6}$) and is solved in a ADMM scheme\footnote{Code ported from \url{https://github.com/yqueau/normal_integration}.}.

As the BRDF samples $j$ (see Equation~\ref{eq:brdf_inv}) depend on the unknown depth, they cannot be directly computed to be input to the network. To overcome this issue, we employ an iterative scheme were the previous estimate of the geometry is used. The procedure involves computing the near to far conversion as described, obtaining a new normal map estimate through the CNN and finally numerical integration. See  Figure~\ref{fig:normal_pred} for an example of intermediate results of our iterative procedure. As it is the case in competing classical methods \cite{logothetis2017semi,queau2018led}, this iterative procedure is initialised with a flat plane at the approximate mean distance.


\begin{figure*}[ht]
\includegraphics[width=\textwidth]{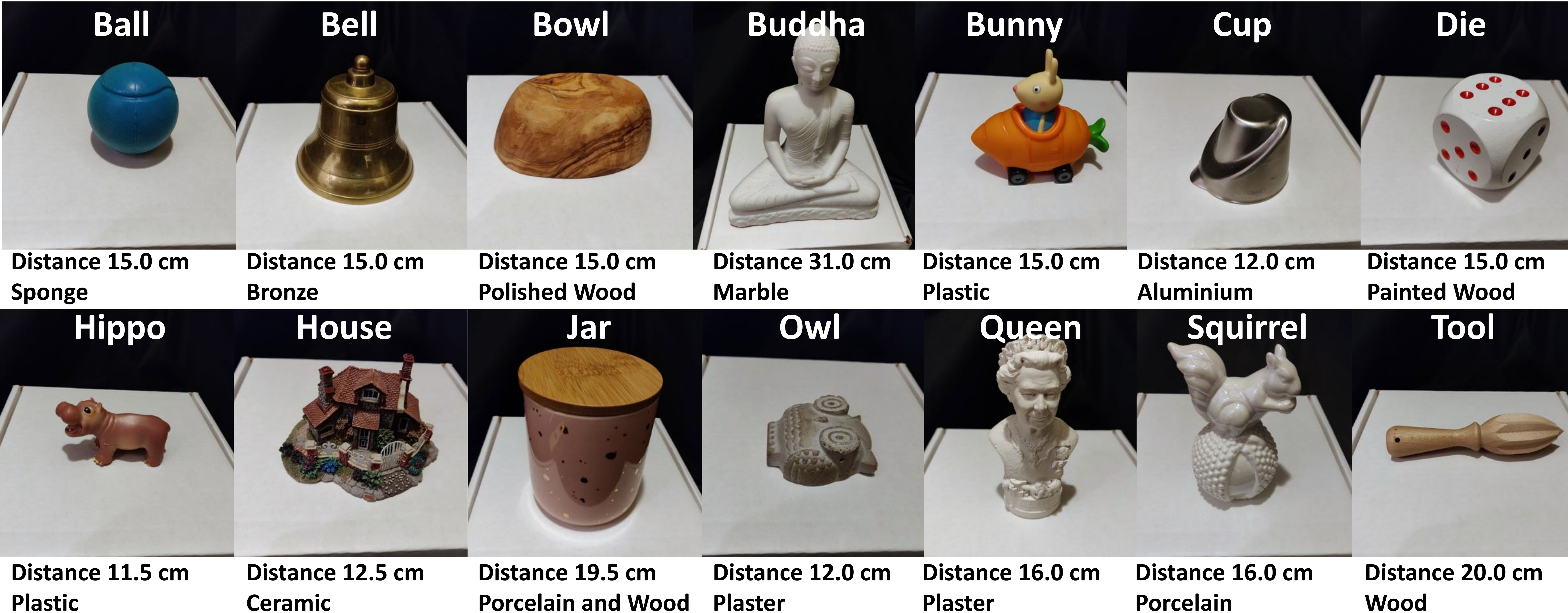}
\caption{Top view of the objects captured for LUCES dataset. Below every object the acquisition distance between the object and the camera, and the material of the object are reported. } 
\label{fig:luces_image_samples}
\end{figure*}

\begin{figure*}[t]
\centering
 \includegraphics[height=0.14\textwidth,trim={0cm 0cm 0cm  0cm},clip]{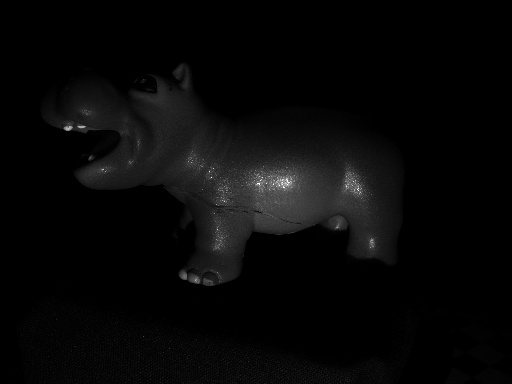}
  \includegraphics[height=0.14\textwidth,trim={0cm 0cm 0cm  0cm},clip]{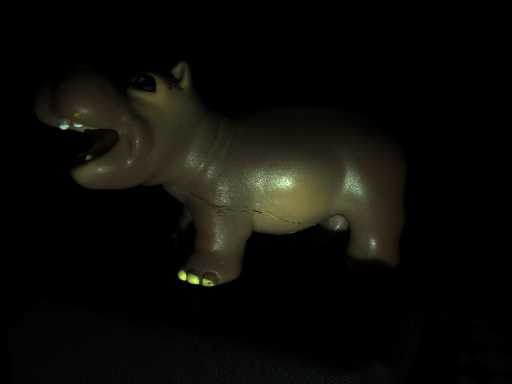}
 \includegraphics[height=0.14\textwidth,trim={1.5cm 0.9cm 1.5cm  0.9cm},clip]{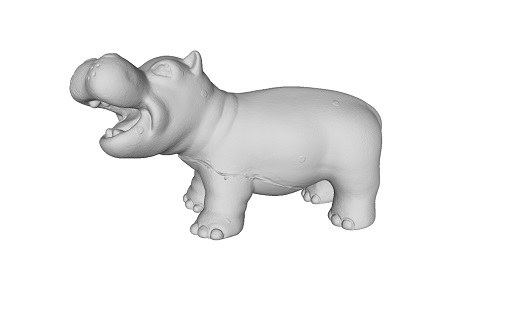}
  \includegraphics[height=0.14\textwidth,trim={0cm 0cm 0cm  0cm},clip]{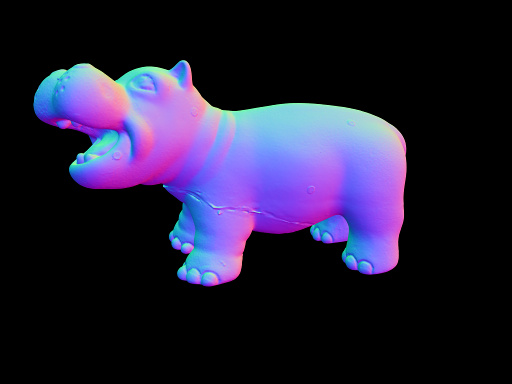}
 \includegraphics[height=0.14\textwidth,trim={0cm 0cm 0cm  0cm},clip]{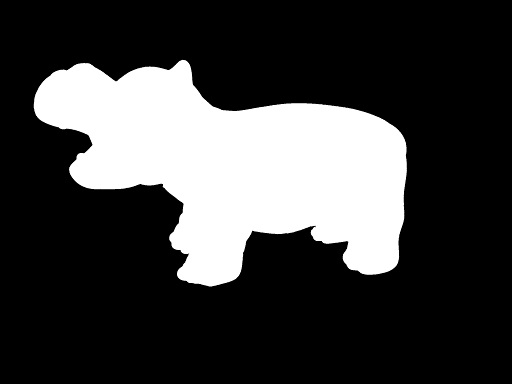}
 \caption{Demonstration of the processing steps performed per object in LUCES dataset. Firstly, compensation for radial distortion and demosaicing is performed on raw images to get RBG ones (left). CT-scanned ground truth meshes are aligned with RGB images and ground truth normal maps are rendered (middle). Segmentation masks are also manually generated.
 }
 \label{fig:steps}
\end{figure*}

\section{LUCES Dataset}
\label{sec:LUCES}

This section gives an overview of the data capture and calibration procedure for the LUCES dataset, first presented in \cite{mecca2021luces}.

\noindent
\textbf{The Photometric stereo setup.} Our setup (see Figure \ref{fig:setup}, left) consists of the following main components:
\begin{itemize}
\setlength\itemsep{0.1mm}
    \item RGB camera FLIR bfs-u3-32s4c-c with 8mm lens,
    \item 52 LED Golden Dragon OSRAM,
    \item variable voltage for adjustable LED power,
    \item Arduino Mega 2560.
\end{itemize}

A custom printed circuit board (PCB) has been designed to host 52 bright LED controlled with by an Arduino Mega. The configuration of the LEDs was planar around the camera. 
A set of 52 images was captured per object. The camera parameters (aperture and shutter speed) and LED voltage were adjusted to achieve the best object exposure, which is very critical for specular objects. All camera prepossessing was turned off during the acquisition, including white-balance and analog gain. 
Several optomechanical tools have been used for holding the camera and the PCB jointly. A manual XYZ translation stage with differential adjusters has been used to positioning the camera accurately through the printed circuit board. In order to limit interreflections and  ambient light, the walls surrounding the setup have been covered with black, polyurethane-coated nylon fabric. 

\noindent
\textbf{Camera intrinsics.} This is performed using 100 checkerboard images and the OpenCV calibration toolbox. Fourth degree radial distortion is estimated and this is used to rectify all the images. The calibration re-projection error was $0.42$px. The RAW data (before demosaicing and rectification) are also available.

\begin{figure*}[ht]
\centering
\includegraphics[width=\textwidth]{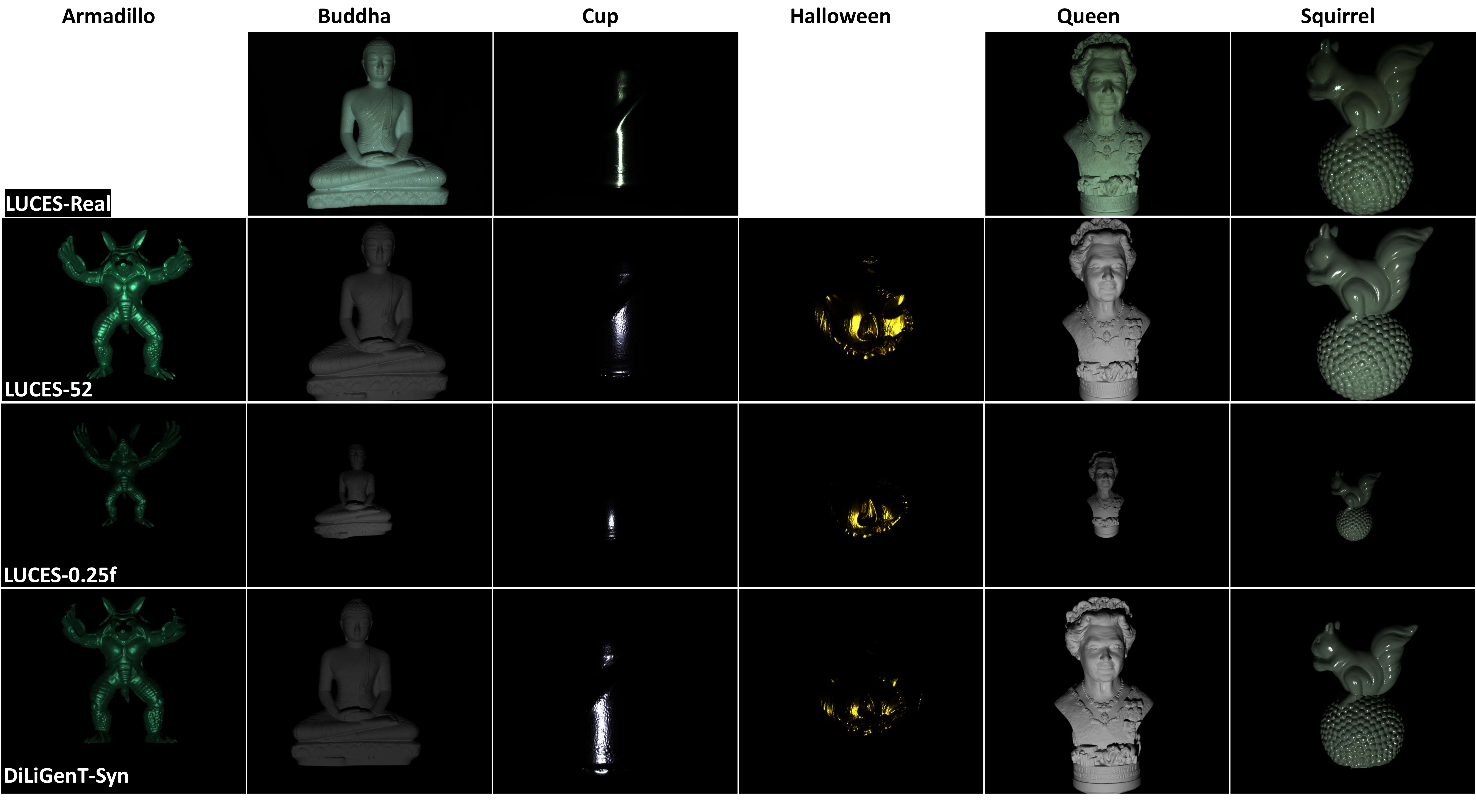}
\caption{One image example for all synthetic objects rendered as well as the corresponding real ones in top row (expect for Armadillo and Halloween that have no real counterparts). The second  row aims to closely match the configuration of LUCES~\cite{mecca2021luces}, the third row reduces the focal length by a factor of 4 and the bottom row aims to closely match DiLiGenT~\cite{diligentshi2016benchmark}.}
\label{fig:synthetic_images}
\end{figure*}

\noindent
\textbf{Light calibration.} This section presents the method used to estimate all the point light parameters  introduced in  Section \ref{sec:method_point} ( $ \{ \mathbf{P}_m, \phi_m, \mathbf{D}_m, \mu _m\} $). To do so, we captured PS images of a purely diffuse reflectance plane i.e. 99\% nominal reflectance in UV-VIS-NIR wavelength range (350 - 1600nm). To have an initial estimate of $\phi_m$, we measured the brightness of the LEDs with a LuxMeter.  For every object, the calibration plane was captured twice, at different distances, in order to get data redundancy and produce a more accurate calibration. Thus, the Lambertian calibration object with albedo $\rho$ and surface normal $\mathbf{N}$, should satisfy the resulting image irradiance equation:
\begin{equation}
\label{eq:irad:lamb}
i_m= \phi_m a_m \rho {\hat{\mathbf{L}}}_m \cdot\hat{\mathbf{N}}.
\end{equation}
The irradiance Equation~\ref{eq:irad:lamb} was implemented into a differentiable renderer (using Keras of Tensorflow v2.0) with the LED parameters being the model weights thus allowing refinement from a reasonable initial estimate. The parameters were initialised as follows: $\phi_m$ from the LuxMeter, $\mathbf{D}_m=[0,0,1]$, $\mu_m=0.5$, $\mathbf{P}_m$ from the schematic of the printed circuit board of the LEDs and $\rho =0.5$. We used $L_1$ loss function for 30 epochs and converged to around 0.005 error i.e 0.5\% of the maximum image intensity.  The complete calibration parameters are included in the dataset. 
 
\noindent
\textbf{3D ground truth capture.}
Initial version of the 3D ground-truth \cite{mecca2021luces} was acquired with the optical 3D scanner GOM ATOS Core 80/135 
with a reported accuracy of 0.03mm (see Figure \ref{fig:setup}). The GOM scanner uses a stereo camera set-up and more than a dozen scans were performed and fused per object.  In order to keep the geometry of the object consistent with the PS data, no spray coating has been used to ease the acquisition. In this work, we provide more accurate ground truth meshes for all the non-diffuse objects\footnote{ Scans of all the objects except Buddha, House, Owl, Queen were obtained with the Zeiss CT scanner M1500/225 kV which provides an accuracy within the order of 9$\mu m$.}.

\noindent
\textbf{Alignment.} The scans of the objects were aligned and merged using MeshLab \cite{CignoniCCDGR08}. Some manual removal of noisy regions was performed and finally Poisson reconstruction was used in order to obtain full continuous surfaces which are both useful for rendering normal maps and for mutual information alignment. As expected, not all parts of the surfaces of all objects have the same amount of noise, especially the metallic objects (Bell, Cup). Meshes were aligned with the photometric stereo images using the mutual information registration filter of MeshLab \cite{CignoniCCDGR08}. This was initialised manually and pixel perfect accuracy was achieved. Using the aligned meshes, ground truth normal maps were rendered (using Blender). In addition, manual segmentation was performed to remove regions where the GT was unreliable (markers on the objects, holes etc). The steps per object are summarised in Figure~\ref{fig:steps}.

\noindent
\textbf{Dataset overview.} For each of the 14 objects (see Figure~\ref{fig:luces_image_samples}), 52 PS images have been acquired using the BayerRG16 RAW format. The total amount of PS images amounts to 728. For all objects, rectified RGB PS images are available. 
We note that color balancing was not performed on the images as this distorts the saturated pixels and ultimately looses information. 
Instead, RGB light source brightness are provided along with the rest of point light source parameters. 
Both normal map and depth ground truth are be provided in order to evaluate the accuracy of near-field PS methods with either case.

\begin{figure}[!t]
\includegraphics[width=\columnwidth]{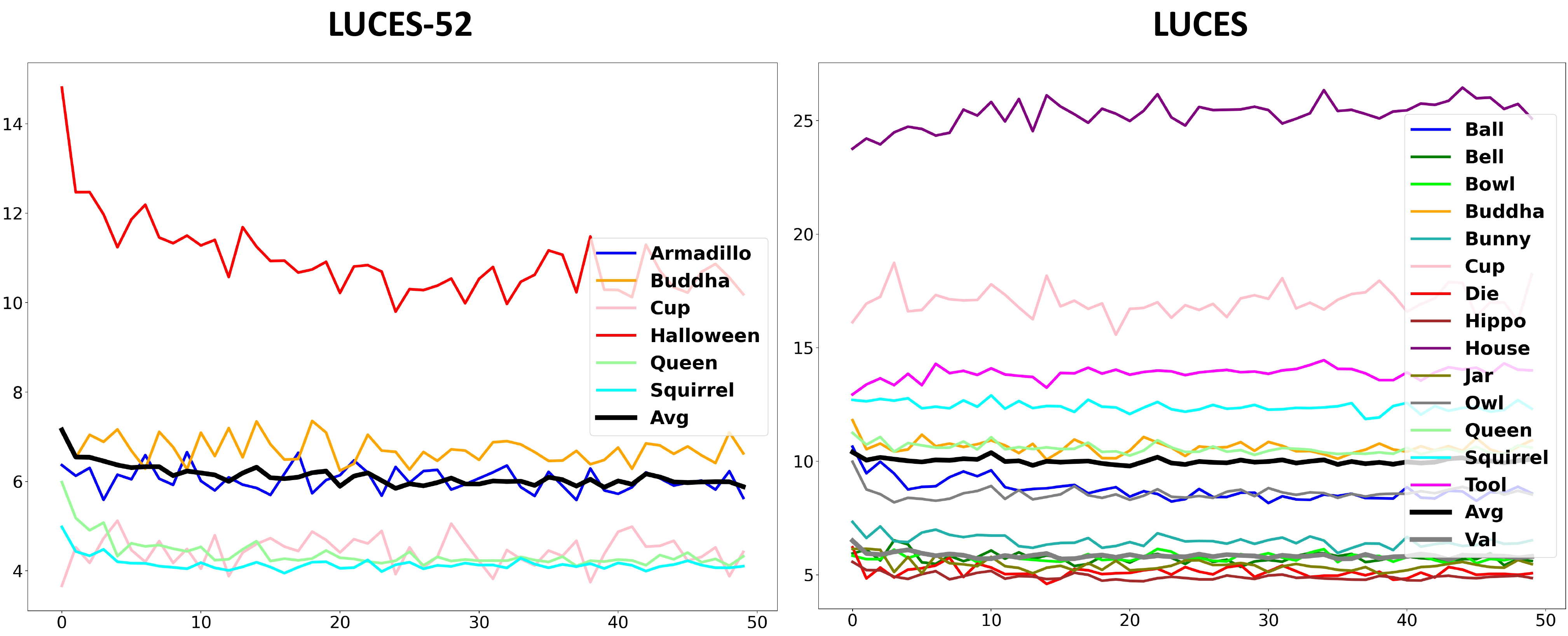}
\caption{MAE evolution (during training) curves illustrate the performance of our setup-agnostic network on predicting normals (NfCNN) for synthetic (left) and real data (right). We note that we used the real DiLiGenT dataset as validation loss and selected the checkpoint (35) were this is minimised. We observe that although the average error is gradually decreasing, for some real objects (House, Cup) the performance is actually getting worse with more training signifying that some real effect is not properly modelled.}
\label{fig:mae_evolution}
\end{figure}

\noindent
\textbf{Non GT objects.} We also captured 15 light images of 3 additional objects shown in Figure~\ref{fig:main_real_qualitative_nogt}.  Metallic-silver Bulldog statue, a porcelain Frog as well as a mutli-object scene featuring a shiny wooden elephant statue in front of the porcelain Squirrel. Bulldog and Frog were too hard to scan and the elephant and squirrel could not be transported in their exact configuration to the CT scanner.


\section{Experimental Setup}
\label{sec:experimental_setup}
In this section we provide various experimental setup details related to CNN training and datasets used for evaluation.

\subsection{CNN Training}  We use the exact architecture of PX-NET \cite{logothetis2021pxnet} which is a miniature version of DenseNet \cite{huang2017densely} with 4.5M parameters. We trained 3 networks, one with general data data and 2 specific ones, one for the LUCES configurations and one for the DiLiGenT one. For the general one, we trained for 50 epochs and selected the checkpoint with best DiLiGenT performance (35). The MAE evolution is shown in Figure~\ref{fig:mae_evolution} for this network. The specific networks converged more quickly taking 9 epochs for LUCES and 25 epochs for DiLiGenT. A batch size of 2400 and 5000 steps per epoch was used in all experiments. It took 20 minutes to complete an epoch on a machine with 2 Titan RTX GPUs.

\subsection{Datasets} We evaluate our method on the real datasets LUCES (see Section~\ref{sec:LUCES}) and DiLiGenT \cite{diligentshi2016benchmark}. Additional evaluation is performed on four synthetic datasets namely  LUCES-52, LUCES-15, LUCES-0.25f and synthetic DiLiGenT. More details about each dataset are provided bellow:

\noindent \textbf{DiLiGenT\cite{diligentshi2016benchmark}.} It contains 96 images of the resolution of $612\times512$px for each of 10 captured objects. It is usually assumed that this is a far field dataset with the directional uniform illumination. However, in reality LEDS were used for the illumination, and their positions $\vect{P}_m$, brightness $\phi _m$ 
and perspective camera parameters are also provided. Using the light positions $\vect{P}_m$, for each object the mean distance can be computed to match the average light directions $\hat{\mathbf{L}}_m$. Finally, LED directions $\mathbf{D}_m$ were assumed perpendicular to the LED plane, and $\mu =0.5$ was also assumed.

\noindent \textbf{Synthetic LUCES-52\cite{mecca2021luces}.} In order to have an estimate of the synthetic to real gap, we chose 4 objects from LUCES and rendered them in exactly the same position and similar materials: Buddha, Queen - Lambertian, Cup - metallic, Squirrel - specular dielectric. We also rendered two additional synthetic objects: Armadillo and Halloween. Armadillo was chosen as it has a challenging geometry (with occlusion boundaries at hands/face) and was rendered with an `intermediate' Disney~\cite{burley2012physically} material (all parameters\footnote{Due to the highly non-linear nature of the BRDF, this material is not necessarily the average brightness.} set to 0.5). The Halloween object was chosen to be rendered metallic-gold as it is very hard to laser scan real metallic objects with concavities and other high frequency surface details. All objects were rendered with the Cycles rendering engine of Blender \cite{blender} to generate realistic global shadows and self reflections. The default  path tracing integrator of Cycles was used using 100 samples per pixel, 8 light bounches as well as no post-processing de-noising.

\noindent \textbf{Synthetic LUCES-15 and LUCES-0.25f}. In addition to the synthetic version of LUCES described above,  we consider another two variations namely Synthetic LUCES-15 and Synthetic LUCES-0.25f. The first one simply contained 15 out of the 52 images per object and was aimed at providing an evaluation at a sparse lighting setting which is usually preferred in practice. LUCES-0.25f was rendered with exactly the same objects and materials by reducing the camera focal length from 8mm to 2mm, in order to simulate a fish-eye lens. Object sizes/positions were also adjusted to keep them aligned in the middle of the field of view. The purpose of this dataset is to test a situation where perspective viewing becomes significant. See Figure~\ref{fig:synthetic_images} for example images.

\noindent \textbf{Synthetic DiLiGenT.} Finally, the same 6 synthetic objects were rendered in the DiLiGenT \cite{diligentshi2016benchmark} configuration. The aim of this experiment was to assess potential performance improvement when the number of lights increases from 52 to 96 and the capture setup becomes for `far-field' - higher distance from the camera and higher focal length.

\subsection{Evaluation Protocol} 

This section describes the evaluation protocol for all the above mentioned datasets.

\noindent \textbf{Competitors.} We compare our method against the far-field CNN approaches of \cite{ikehata2018cnn}, \cite{logothetis2021pxnet}, the near-field variational optimisation methods \cite{logothetis2017semi}, \cite{queau2018led} as well as the recent near-field CNN-based method of \cite{santo2020deep}. For all 3 CNN-based methods, the same network checkpoint was used as the one in the corresponding papers. For all methods, test code was available online\footnote{\small{ \cite{logothetis2017semi}~\url{https://github.com/fotlogo/semi_calib_ps_cvpr2017} \\  \cite{queau2018led} \url{https://github.com/yqueau/near_ps}\\
\cite{santo2020deep}~\url{https://github.com/hiroaki-santo/deep-near-light-photometric-stereo}\\
\cite{ikehata2018cnn}~\url{https://github.com/satoshi-ikehata/CNN-PS}}}. Finally, for the comparison on the DiLiGenT\cite{diligentshi2016benchmark} benchmark (see Table~\ref{tab:diligent}), we also report the numbers of some other competing approaches. 

\noindent \textbf{Naive vs Compensated usage of far-field methods.} In order to demonstrate the importance of our point-light compensation procedure, we compare the usage of the far-field CNN approaches of \cite{ikehata2018cnn}, \cite{logothetis2021pxnet} with/without using it. Naive usage refers to using the raw image values (i.e. $i_m$) without attenuation compensation for computing observation maps as well as the average light direction for each LED. The predicted normals are also integrated using our method in order to have a qualitative shape comparison.

\noindent \textbf{Evaluation metrics}. For most experiments, the evaluation metrics are mean angular error (MAE) on normals (in degrees) as well as mean depth error (in mm) on computed depth maps. We note that the real DiLiGenT \cite{diligentshi2016benchmark} benchmark only reports ground truth normals and also for 3 of our objects, no ground truth was available so the comparison is only qualitative (see Figure~\ref{fig:main_real_qualitative_nogt}). In addition, we note that the variational optimisation methods  \cite{logothetis2017semi}, \cite{queau2018led} only output depth maps, therefore in order to have comparison in the normal domain for them, normal maps are generated with numerical differentiation. Therefore, for the rest of the methods we report 2 types of normal maps namely NfCNN (normals from CNN-network predictions) and NfS  (normals from shape-obtained though numerical differentiation).

\noindent \textbf{Input resolution}. Most LUCES experiments (both real and synthetic), were performed at a quarter resolution i.e. $512\times384$px in order to have a fair comparison with \cite{santo2020deep} which is GPU memory limited (even on their original paper the authors report unable to run on more than $600\times600$px resolution on a 48GB GPU RAM). However, it has to be noted that we are unsure if some of their respective hyperparameters is resolution dependent. For the real LUCES data, we also present our evaluation on full resolution ($2048\times1536$px) images and show that our method is resolution independent (with around $0.1mm$, $0.1^o$  difference between quarter and full scale). DiLiGenT (both synthetic and real) on the other hand offers maximum resolution of $612\times 512$px which was used for all the relevant experiments reported.

\section{Experiments}
\label{sec:experiments}

\begin{figure*}[t]
\centering
\includegraphics[width=\textwidth]{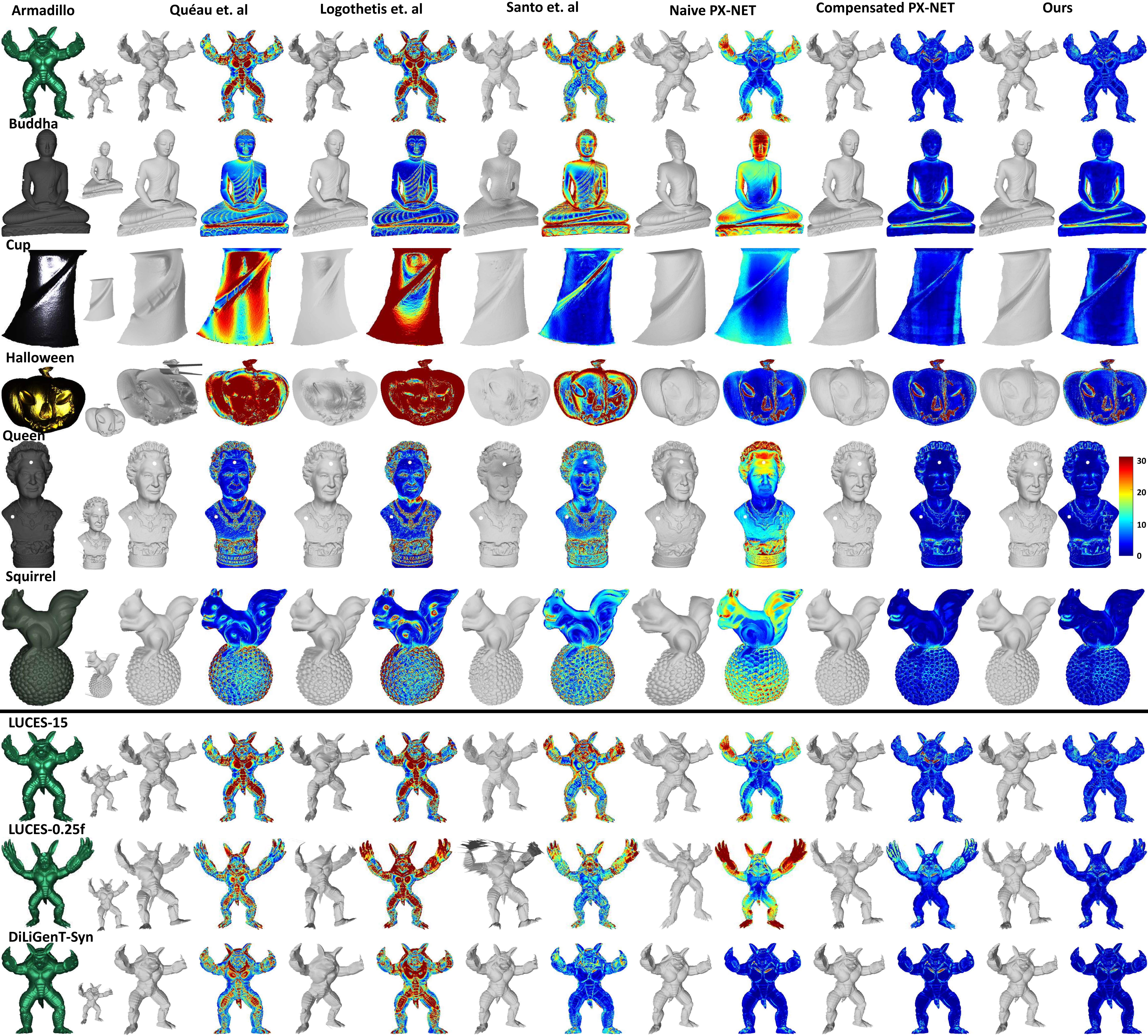}
\caption{Visualisation of the results of Table~\ref{tab:synthetic_all} showing comparison with Logothetis et. al~\cite{logothetis2017semi} and Qu{\'e}au et. al~\cite{queau2018led}, Santo et. al~\cite{santo2020deep} and PX-NET~\cite{logothetis2021pxnet} on all synthetic experiments. All 6 objects of synthetic LUCES-52 are shown and Armadillo is also shown for the other 3 synthetic datasets. For all objects, average PS image and GT depth shape is shown at the left. \cite{logothetis2017semi}, \cite{queau2018led} have very high error on the metallic objects (Cup, Halloween) as well as specular highlights on Squirrel and Armadillo. The naive far-field method PX-NET also has significant global deformation as it does not model the light attenuation effect. In contrast, the compensated version performs very well everywhere except the hands of Armadillo in the LUCES-0.25f due to its training data being rendered without perspective viewing. For \cite{santo2020deep}, the error is concentrated towards the edges of each object as perspective projection is not modelled (this is especially evident on the LUCES-0.25f). The proposed approach achieves best performance in all objects.}
\label{fig:main_synthetic_qualitative}
\end{figure*}

In this section we present the results of the various experiments on the synthetic and real datasets introduced in the previous section. 

\subsection{Synthetic Experiments}

\begin{table*}[!t]
\includegraphics[width=\textwidth]{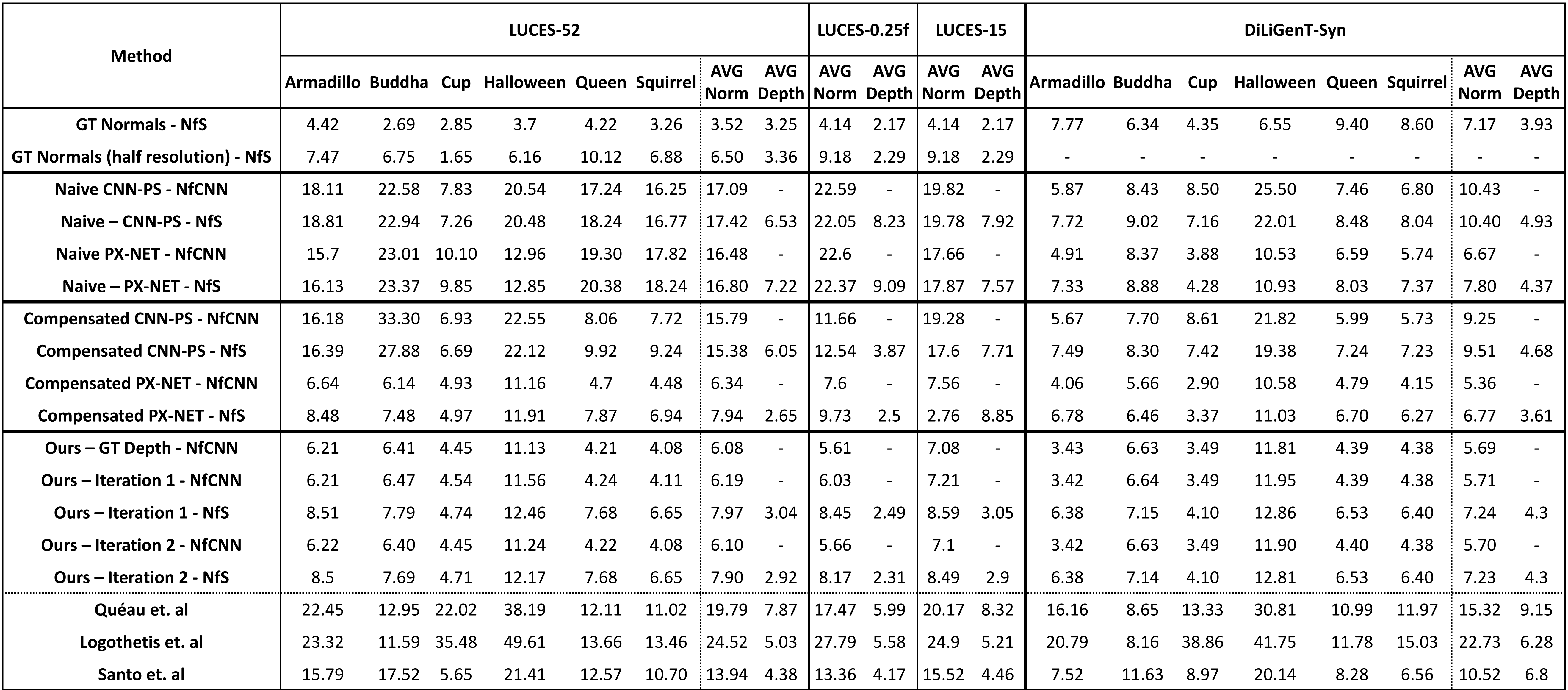}
\caption{Full quantitative comparison on synthetic data. For out method, we report raw normal prediction error as NfCNN and numerically differentiated normals as NfS. Also report NfCNN error when the GT depth is used as initialisation and also NfS error for integrating the GT normals. We compare Logothetis et. al~\cite{logothetis2017semi} and Qu{\'e}au et. al~\cite{queau2018led}, Santo et. al~\cite{santo2020deep}, CNN-PS~\cite{ikehata2018cnn} and PX-NET~\cite{logothetis2021pxnet}. For LUCES-52 and synthetic DiLiGenT, metrics on all objects are show; for the other 2 only average errors are reported (the aim is to observe drop of performance by reducing the number of lights or focal length). Note that for the NfCNN-GT  reported metric, only the normal error is meaningful.}
\label{tab:synthetic_all}
\end{table*}

This section explains the synthetic experiments which are summarised in Table~\ref{tab:synthetic_all}. A further breakdown per category follows.

\noindent \textbf{Shape integration.} The first experiment we conducted aimed at calibrating the quality of the numerical integration of the normal map. As no realistic depth map is C2 continuous, GT normals are not compatible with the GT depth. Indeed, integrating the GT normals and then re-calculating them with numerical differentiation introduces $3.52^o$ MAE on average (Table \ref{tab:synthetic_all} top) on full resolution and even reaches $6.50^o$ at half resolution as the naive numerical differentiation is very resolution dependent. Therefore, we do not compare raw network predictions (NfCNN) and MAE after differentiation of the surface (NfS) and expect the first figure to be lower for networks trained to regress normals\footnote{Some more sophisticated shape differentiation method such as \cite{smith2020morphable} would probably reduce the discrepancy between these 2 metrics but that would not alter the discussion of this section.}.

\noindent \textbf{Naive usage of far-field networks.} 
The next experiment consists of naively using the far-field methods \cite{ikehata2018cnn, logothetis2021pxnet} with no point light compensation. As expected, in all 'near field' results (except in synthetic DiLiGenT), both normal and depth errors are significantly higher than all other competitors and this is better understood visually in Figure~\ref{fig:main_synthetic_qualitative}; the shape is 'locally correct' with  no bumps at specular highlights or other similar artifacts but still severely distorted. To add to this point, using these network as part of our iterative process (marked as Compensated) drastically improves the result and in fact they outperform the variational optimisation methods of \cite{logothetis2017semi} and \cite{queau2018led}. The difference between naive and compensated is significantly lower on the synthetic DiLiGenT as scale of this dataset is mostly far-field with the point light effect being less important. 

\begin{figure*}[t]
\centering
\includegraphics[width=\textwidth]{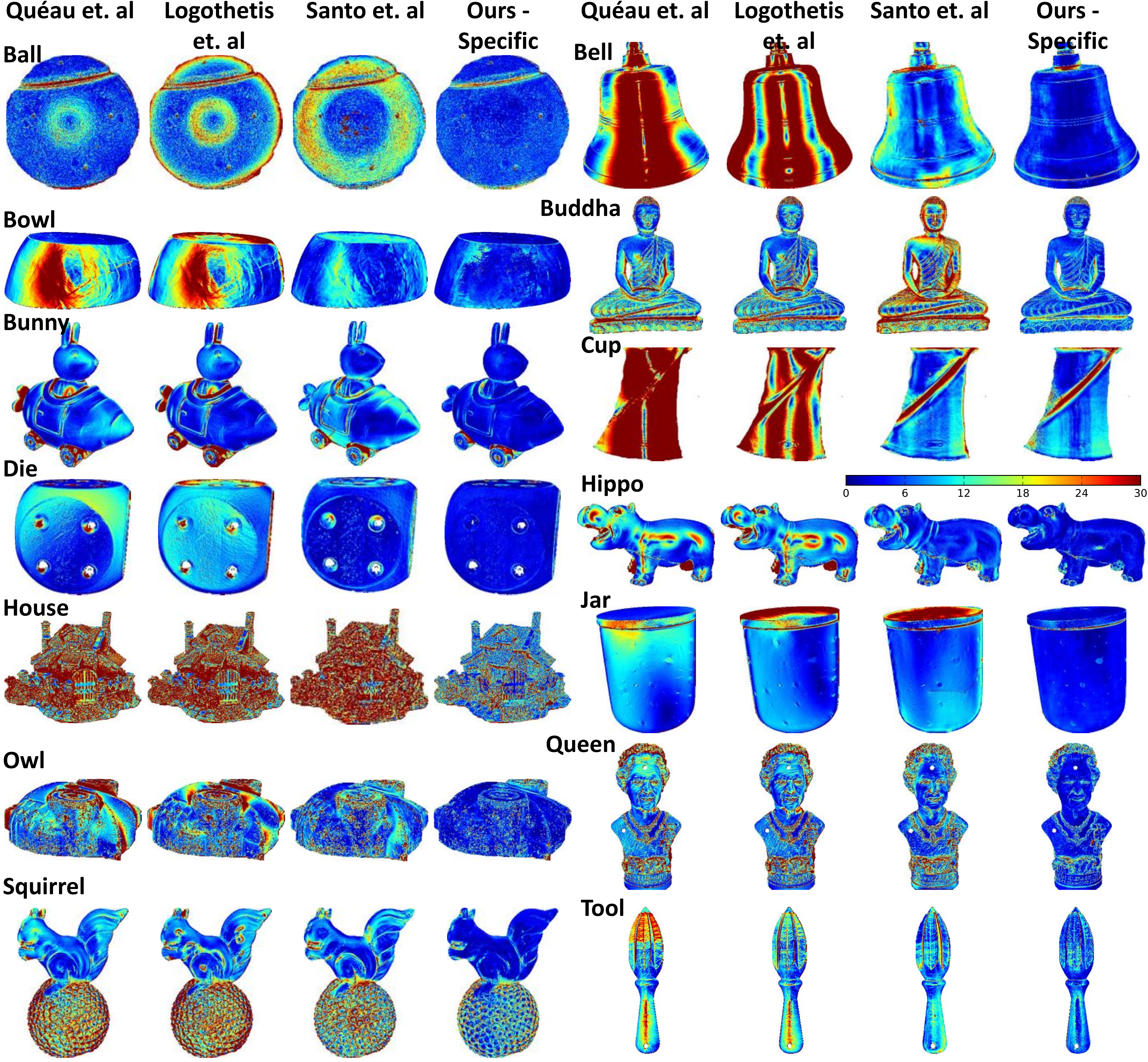}
\vspace{-0.25cm}
\caption{Visualisation of the normal error map comparison for all objects of real LUCES (Table~\ref{tab:luces}) and all near-field methods: Logothetis et. al \cite{logothetis2017semi},  Qu{\'e}au et. al \cite{queau2018led}, Santo et. al \cite{santo2020deep}.}
\label{fig:real_luces_normals}
\end{figure*}

\noindent \textbf{Proposed method}. We report results of our method using the setup agnostic network. We report error at iteration 1 (i.e. compensation using the initial planar geometry estimate) and iteration 2 and observe a marginal improvement signifying that the process has converged. We also report NfCNN where the point light compensation has been performed with the GT depth to estimate the limiting performance of the iterative method. We again confirm that this is only marginally better than iteration 2 error confirming that our network is not very sensitive to depth initialisation.

\noindent \textbf{Material variation}. We note that the proposed method performs similarly in all 6 objects despite the significant material variations. This is not the case for the variational optimisation competitors (\cite{logothetis2017semi}, \cite{queau2018led}) which are significantly worse on the metallic objects (Cup, Halloween) than the Lambertian ones (Buddha, Queen). Quite surprisingly, \cite{santo2020deep} performs worse on the Lambertian objects possibly signifying lack of training data with exact Lambertian reflections.

\noindent \textbf{LUCES-0.25f.} We observe no drop of performance between LUCES-52 and its lower focal length counter part verifying that our method correctly compensates for the effect of perspective viewing. In contrast, for the naive far-field methods the error is increased. 

\begin{figure*}[t]
\centering
\includegraphics[width=\textwidth]{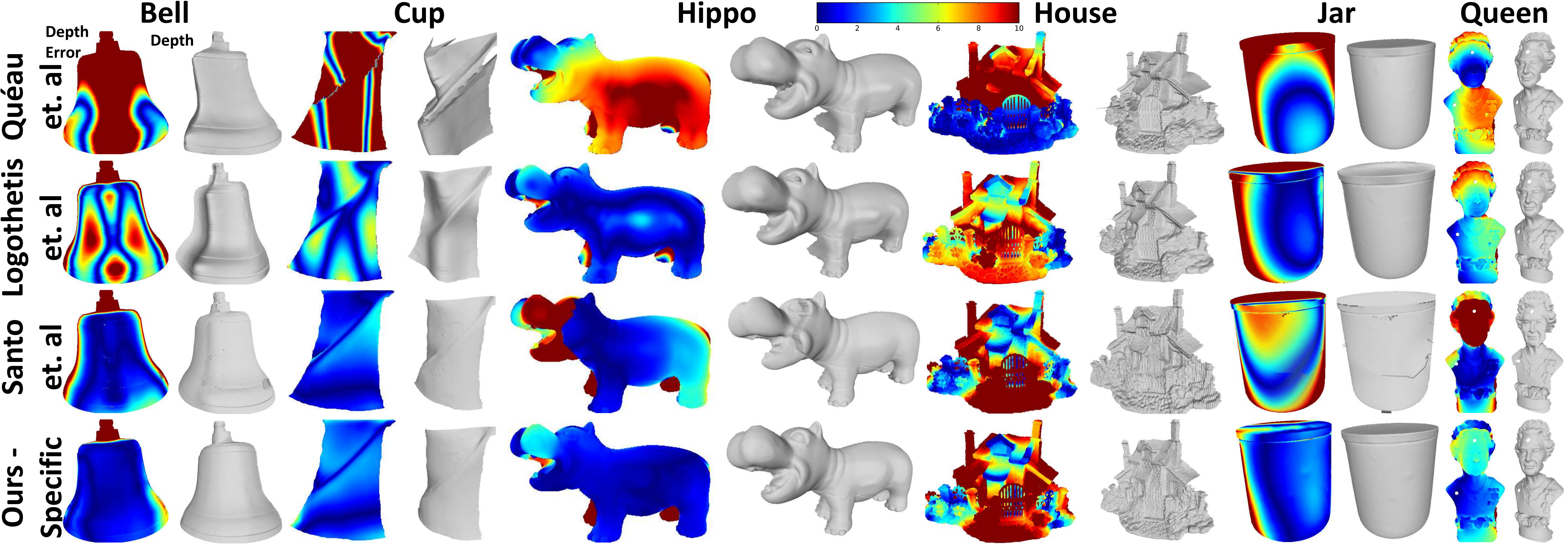}
\caption{Output surface comparison for 6 objects of real LUCES (see Table~\ref{tab:luces} for quantitative results) and methods of Logothetis et. al~\cite{logothetis2017semi},  Qu{\'e}au et. al~\cite{queau2018led}, Santo et. al~\cite{santo2020deep}. This is shown qualitatively through the 3D meshes and well as depth error maps (errors in mm).}
\label{fig:real_luces_shape}
\end{figure*}

\begin{table*}[t]
\setlength{\tabcolsep}{3.0pt} 
\begin{center}
\begin{footnotesize}
\resizebox{1.0\columnwidth}{!}{%
\begin{tabular}{ | c | c c c c c c c c c c c c c c c | c|}
\hline
\textbf{Method} & \textbf{Error} & \textbf{Ball} & \textbf{Bell} & \textbf{Bowl} & \textbf{Buddha} & \textbf{Bunny} & \textbf{Cup} & \textbf{Die} & \textbf{Hippo} & \textbf{House} & \textbf{Jar} & \textbf{Owl} & \textbf{Queen} & \textbf{Squirrel} & \textbf{Tool} & \textbf{Average} \\ \hline
\multirow{3}{*}{\vspace{0.4cm} \textbf{Logothetis et. al \cite{logothetis2017semi}}} & MAE & 12.55  & 36.02  & 16.66  & 14.59  & 13.01  & 28.08  & 11.54  & 14.27  & 32.20  & 12.21  & 16.61  & 15.99  & 17.73  & 15.72  & 18.37  \\ 
& MZE & 1.60  & 5.60  & 6.11  & 5.49  & 2.14  & 3.17  & 4.35  & 2.09  & 7.62  & 6.08  & 4.06  & 5.11  & 2.10  & 5.34  & 4.35   
 \\  \hline
\multirow{3}{*}{\vspace{0.4cm} \textbf{ Qu{\'e}au et. al \cite{queau2018led}}} & MAE & 8.96  & 28.16  & 16.16  & 15.67  & 12.60  & 43.20  & 9.49  & 13.93  & 33.01  & 10.30  & 17.39  & 16.01  & 16.55  & 16.86  & 18.45  \\ 
& MZE & 3.67  & 12.83  & 6.54  & 7.72  & 2.56  & 13.03   & 4.14  & 8.21  & 6.19  & 11.97  & 5.51  & 5.92  & 4.67  & 6.93  & 7.14  \\   \hline
\multirow{3}{*}{\vspace{0.4cm} \textbf{Santo et. al \cite{santo2020deep}}} & MAE & 13.27  & 10.03  & 7.27  & 19.22  & 9.44  & 11.74  & 4.95  & 7.20  & 31.49  & 10.00  & 12.09  & 13.43  & 13.24  & 10.90  & 12.45  \\ 
& MZE & 2.48  & 2.55  & 3.94  & 6.24  & 5.13  & 1.44  & 2.65  & 3.76  & 8.13  & 7.11  & 5.06  & 5.47  & 2.07  & 5.96  & 4.43  \\ \hline  \hline
\multirow{3}{*}{\vspace{0.4cm} \textbf{Naive PX-NET~\cite{logothetis2021pxnet}}} & MAE & 14.23  & 13.47  & 8.54  & 29.99  & 18.99  & 13.84  & 15.59  & 21.17  & 35.23  & 19.38  & 21.19  & 21.77  & 25.63  & 19.33  & 19.88  \\ 
& MZE & 4.16  & 4.37  & 7.20  & 18.22  & 3.36  & 1.36  & 5.25  & 5.46  & 8.74  & 11.15  & 2.88  & 5.56  & 3.76  & 3.83  & 6.09\\ \hline 
\multirow{3}{*}{\vspace{0.4cm} \textbf{US - Compensated}} & MAE & 10.90  & 8.99  & 6.03  & 12.27  & 7.77  & 14.46  & 8.02  & 11.92  & 29.80  & 7.00  & 17.69  & 11.25  & 13.87  & 13.44  & 12.39 \vspace{0.05cm} \\ 
\textbf{PX-NET} & MZE & 0.80  & 2.03  & 2.16  & 3.80  & 2.55  & 1.60  & 1.81  & 3.30  & 8.79  & 4.44  & 4.60  & 3.64  & 1.54  & 2.17  & 3.09   \\ \hline \hline
\multirow{3}{*}{\vspace{0.4cm} \textbf{US - Specific}} & MAE & 8.62  & 6.36  & 6.41  & 11.59  & 6.85  & 14.09  & 4.76  & 5.67  & 22.53  & 5.63  & 8.86  & 9.82  & 11.84  & 11.19  & 9.59  \\ 
& MZE & 0.59  & 1.66  & 2.22  & 3.54  & 2.39  & 1.84  & 1.55  & 1.51  & 8.34  & 2.32  & 4.23  & 3.12  & 1.15  & 5.12  & 2.83  \\  \hline
\multirow{3}{*}{\vspace{0.4cm} \textbf{US - General}} & MAE & 8.86  & 7.51  & 5.96  & 11.60  & 7.07  & 15.27  & 5.19  & 5.54  & 22.91  & 6.14  & 8.86  & 9.96  & 11.77  & 11.56  & 9.87  \\ 
& MZE & 0.58  & 1.72  & 2.07  & 3.82  & 2.33  & 2.16  & 1.86  & 1.85  & 8.84  & 2.71  & 4.28  & 2.85  & 0.76  & 3.67  & 2.82   \\ 
 \hline
 \multirow{3}{*}{\vspace{0.4cm} \textbf{US - General Full res.} } & MAE & 8.84  & 7.51  & 5.95  & 11.59  & 7.06  & 15.35  & 5.19  & 5.60  & 22.97  & 6.19  & 8.89  & 9.97  & 11.77  & 11.64  & 9.90  \\
& MZE & 0.58  & 1.77  & 1.96  & 3.74  & 2.32  & 2.17  & 1.89  & 1.83  & 8.86 & 2.63  & 4.23  & 2.81  & 0.76  & 3.69  & 2.80 \\ 
 \hline \hline
\end{tabular}
}
\end{footnotesize}
\end{center}
\caption{Evaluation on the LUCES~\cite{mecca2021luces} benchmark compering with Logothetis et. al \cite{logothetis2017semi},  Qu{\'e}au et. al \cite{queau2018led}, Santo et. al \cite{santo2020deep}, and PX-NET~\cite{logothetis2021pxnet}. Mean angular error (MAE in degrees) of predicted normals and mean depth errors (MZE in mm) are reported. }
\label{tab:luces}
\end{table*}

\noindent \textbf{LUCES-15.} We observe small drop of performance in the normal error between LUCES-52 and the 15 lights version ($7.1^o$ to $6.1^o$) but the depth error remains practically the same - $2.9mm$. This is probably explained by the fact that in the low light setting a few points become unsolvable (inflating the mean error) but the overall surface can still be recovered. This is not the case for \cite{santo2020deep} were both normal and depth errors increase. The variational optimisation competitors (\cite{logothetis2017semi} and \cite{queau2018led}) also have minimal drops of performance in this low light setting. A surprising result is that compensated PX-NET is also performing similarly between the 52 and 15 lights settings even though it was trained with a minimum of 50 lights. This is probably explained by the fact that it was trained to be very resilient to shadows which essentially reduce the amount of active lights.


\subsection{Real Data evaluation.} This section presents the results of the real data evaluation on the LUCES~\cite{mecca2021luces} and DiLiGenT \cite{diligentshi2016benchmark} benchmarks.

\begin{figure*}[t]
\centering
\includegraphics[width=\textwidth]{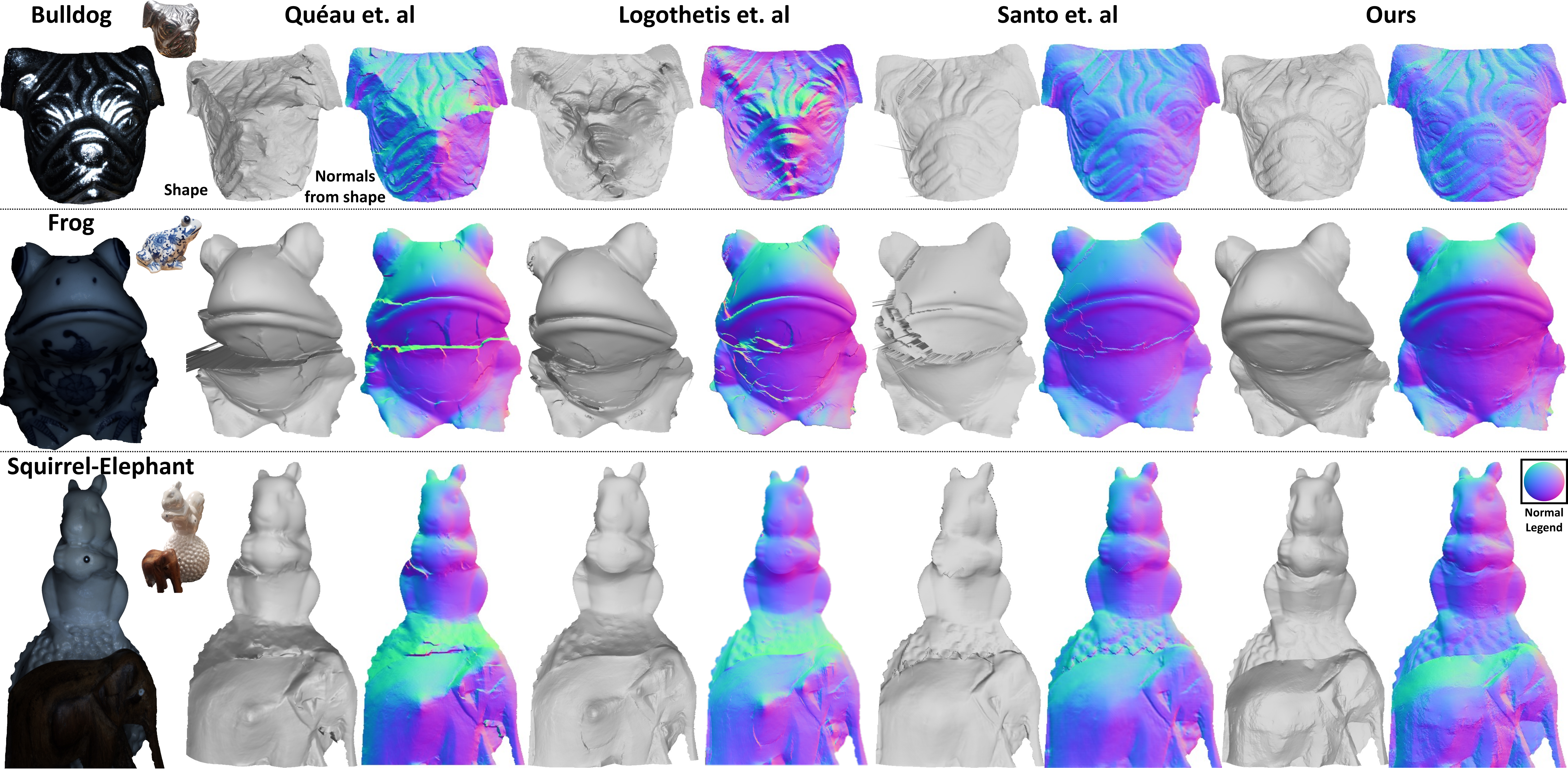}
\vspace{-0.25cm}
\caption{Qualitative comparison of the proposed method with~\cite{queau2018led},~\cite{logothetis2017semi} and~\cite{santo2020deep}. The first column shows the average Photometric Stereo image. In contrast to competition, the proposed approach has no visible deformation on the metallic object or the specular highlight in the middle of the elephant. In addition, there is a smooth recovery of belly of the Frog despite the shadows, as well as the bottom of Squirrel despite self reflection.}
\label{fig:main_real_qualitative_nogt}
\end{figure*}

\noindent \textbf{LUCES \cite{mecca2021luces}}. Table~\ref{tab:luces} shows the quantitative evaluation on LUCES with qualitative comparison through normal maps in Figure~\ref{fig:real_luces_normals} and shapes in Figures~\ref{fig:real_luces_shape} and \ref{fig:main_real_qualitative_nogt}. We achieve the best performance in all objects with the exception of the metallic Cup 
were \cite{santo2020deep} is the best performer.  This may be due to the use of a patch-based network which is able to extract the more information from noisy metallic data. Finally, on the qualitative only data of Figure~ \ref{fig:main_real_qualitative_nogt}, we note that optimisation competitors (\cite{logothetis2017semi} and \cite{queau2018led}) struggle at the metallic Bulldog. This is not the case for \cite{santo2020deep} which seems to struggle at the high curvature region of the Frog neck. The proposed method performs reasonably on all 3 objects.

\noindent \textbf{Synthetic to real gap}. We observe that we perform significantly worse on the real LUCES objects with respect to their synthetic counterparts. As the geometry and lights are similarly matched we conclude that more research is needed in modeling and sampling realistic materials as well as other potential corruptions of real images (better noise models). This is most evident for the metallic Cup were the normal error increases from  $4.5^o$to $14.1^o$. However, for all CNN-based methods (ours, \cite{logothetis2021pxnet,santo2020deep}) the material's specularity does not seem be a significant factor of performance. Indeed, convex regions (where self reflections are negligible) are consistently recovered correctly regardless of the material: diffuse head of Queen, bronze Bell, plastic Hippo, wooden Bowl; with the only exception being the aluminium Cup. This is a clear advantage of CNN methods against the classical ones that require diffuse or mostly diffuse materials.

\noindent \textbf{Normal vs depth errors}. We notice that the normal predictions are more noisy as opposed to depth predictions. This could be due to noisy estimates of the normals from the ground truth meshes which is inevitable for any laser scanner (see in particular the Ball in Figure \ref{fig:real_luces_normals}). As the ground truth depth is more reliable, it is a better evaluation metric compared to the `ground truth' normals.  See Figure \ref{fig:real_luces_shape} for depth evaluation.  

\noindent \textbf{Error distribution}. We observe that the hardest regions are the ones containing high frequency details (sharp boundaries) such as House, bottom part of the Squirrel, details of the Queen, etc. An interesting observation is also that for \cite{santo2020deep} there is growing inaccuracy towards the external part of the reconstruction (see Bell, Cup and Jar in Figure \ref{fig:real_luces_normals}) which is probably due to the orthographic camera assumption.

\noindent \textbf{DiLiGenT \cite{diligentshi2016benchmark}}. Final evaluation at Table~\ref{tab:diligent}. We note that even though this dataset is usually considered far-field with directional lights, our point light compensation procedure improves the performance of PX-NET~\cite{logothetis2021pxnet} (the best performing far-field method) from $6.28^o$ to $5.85^o$ demonstrating the importance of point-light modeling in real data. It is also interesting and somewhat surprising that compensated PX-NET also outperforms our general network ($5.85^o$ vs $5.89^o$) and that signifies that perspective viewing (which is the most important difference of these networks) is not significant on this dataset as opposed to LUCES, as shown in Table~\ref{tab:luces}. Finally, we note that the best performing method is the DiLiGenT-specific network which is not really surprising even though the margin is quite small ($5.66^o$ vs $5.85^o$).

\noindent \textbf{Specific setup network.} Finally we note that setup specific networks are marginally better than the setup agnostic one (which took more time to converge though) signifying that the light distribution is not a big challenge for the CNN.


\begin{table}[t]
\setlength{\tabcolsep}{3.0pt} 
\begin{tabular}{ | c | c c c c c c c c c c|c|}
\hline
\textbf{Method} & \textbf{Ball} & \textbf{Bear} & \textbf{Buddha} & \textbf{Cat} & \textbf{Cow} & \textbf{Goblet} & \textbf{Harvest} & \textbf{Pot1} & \textbf{Pot2} & \textbf{Reading} & \textbf{AVG}  \\ \hline
\textbf{SPLINE-Net}\cite{zheng2019spline} & 1.74 & 4.65 & 9.14 & 5.48 & 9.55 & 9.38 & 24.44 & 5.91 & 7.9 & 12.77 & 9.1   \\ 
\textbf{ICML} \cite{taniai2018neural} & 1.47 & 5.79 & 10.36 & 5.44 & 6.32 & 11.47 & 22.59 & 6.09 & 7.76 & 11.03 & 8.83   \\ \textbf{Exemplars} \cite{hui2016shape} & 1.33 & 5.58 & 8.48 & 4.88 & 8.23 & 7.57 & 15.81 & 5.16 & 6.41 & 12.08 & 7.55   \\ 
\textbf{PS-FCN} \cite{chen2020deep} & 2.7 & 4.8 & 6.2 & 7.7 & 7.2 & 7.5 & 7.8 & 10.9 & 6.7 & 12.4 & 7.4  \\ 
\textbf{CNN-PS}\cite{ikehata2018cnn} & 2.2 & 4.1 & 7.9 & 4.6 & 8 & 7.3 & 14 & 5.4 & 6 & 12.6 & 7.21   \\ 
\textbf{Inverse Model} \cite{wang2020non} & 1.78 & 4.12 & 6.09 & 4.66 & 6.33 & 7.22 & 13.34 & 6.46 & 6.45 & 10.05 & 6.65 \\ 
\textbf{PX-NET} \cite{logothetis2021pxnet} & 2.0 & 3.6 & 7.6 &4.4 & 4.7 & 6.9 & 13.1 & 5.1 & 5.1 & 10.3 & 6.28 \\ 
\textbf{Santo et. al} \cite{santo2020deep} & 5.8 & 5.87 & 9.92 & 6.44 & 7.75 & 8.66 & 15.68  & 7.84  & 9.86 & 13.87 & 9.17 \\ \hline
\textbf{Ours - Comp. PX-NET}  & 1.83 & 3.04 & 6.95 & 3.26 & 4.86 & 5.96& 13.77 & 4.03 & 4.78 & 10.00 & 5.85\\
\textbf{Ours - General}  & 2.62  & 2.87  & 6.85  & 3.52  & 5.98  & 5.97  & 12.24  & 4.20  & 5.24  & 9.39  & 5.89 \\
\textbf{Ours - Specific}  & 1.60  & 2.96  & 7.34  & 3.32  & 4.74  & 5.72  & 12.42  & 4.18  & 5.00  & 9.34  & 5.66 \\
\hline
\end{tabular} 
\caption{Quantitative comparison of the proposed method on the DiLiGenT benchmark \cite{diligentshi2016benchmark}. The competitors are the far-field methods: SPLINE-Net\cite{zheng2019spline}, ICML \cite{taniai2018neural}, PS-FCN \cite{chen2020deep},CNN-PS\cite{ikehata2018cnn}, Inverse Model \cite{wang2020non} and PX-NET \cite{logothetis2021pxnet} as well as the near-field method by Santo et. al \cite{santo2020deep}.}

\label{tab:diligent}
\end{table}

\section{Conclusion}
\label{sec:conclusion}
In this work we presented a CNN-based approach tackling the point light Photometric Stereo problem in both near and far-field setting. We leveraged the capability of CNNs to learn to predict surface normals from reflectance samples for a wide variety of materials and under global illumination effects such as such as shadows and interreflection. Numerical integration is used to compute the depth from predicted normals and this in turn is used to compensate the input images to compute reflectance samples for the next iteration. Finally, in order to measure the performance of our approach for near-field point-light source PS data, we introduced the LUCES dataset containing 14 objects imaged in a configuration where attenuation due to point lights sources and perspective viewing are significant.

\FloatBarrier
\FloatBarrier
%

\bibliographystyle{spmpsci}      

\bibliography{egbib}

\end{document}